\documentclass{article}

\usepackage{PRIMEarxiv}

\usepackage[utf8]{inputenc} % allow utf-8 input
\usepackage[T1]{fontenc}    % use 8-bit T1 fonts
\usepackage{hyperref}       % hyperlinks
\usepackage{url}            % simple URL typesetting
\usepackage{booktabs}       % professional-quality tables
\usepackage{amsfonts}       % blackboard math symbols
\usepackage{nicefrac}       % compact symbols for 1/2, etc.
\usepackage{microtype}      % microtypography
\usepackage{lipsum}
\usepackage{fancyhdr}       % header
\usepackage{graphicx}       % graphics
\usepackage{amsmath}
\usepackage{subfig}
\usepackage{graphicx}
\graphicspath{{media/}}     % organize your images and other figures under media/ folder
\usepackage{multirow}

\newcommand\Tau{\mathcal{T}}

\title{StorSeismic: A new paradigm in deep learning for seismic processing
}

\author{
  Randy Harsuko and Tariq Alkhalifah \\
  King Abdullah University of Science and Technology \\
  Thuwal, Kingdom of Saudi Arabia \\
  \texttt{mochammad.randycaesario@kaust.edu.sa} \\
}

\begin{document}
\maketitle

\begin{abstract}
    Machine learned tasks on seismic data are often trained sequentially and separately, even though they utilize the same features (i.e. geometrical) of the data. We present StorSeismic, as a framework for seismic data processing, which consists of neural network pre-training and fine-tuning procedures. We, specifically, utilize a neural network as a preprocessing model to store seismic data features of a particular dataset for any downstream tasks. After pre-training, the resulting model can be utilized later, through a fine-tuning procedure, to perform tasks using limited additional training. Used often in Natural Language Processing (NLP) and lately in vision tasks, BERT (Bidirectional Encoder Representations from Transformer), a form of a Transformer model, provides an optimal platform for this framework. The attention mechanism of BERT, applied here on a sequence of traces within the shot gather, is able to capture and store key geometrical features of the seismic data. We pre-train StorSeismic on field data, along with synthetically generated ones, in the self-supervised step. Then, we use the labeled synthetic data to fine-tune the pre-trained network in a supervised fashion to perform various seismic processing tasks, like denoising, velocity estimation, first arrival picking, and NMO. Finally, the fine-tuned model is used to obtain satisfactory inference results on the field data.

\end{abstract}

% keywords can be removed
\keywords{Seismic processing \and Machine learning \and Transformer \and Self-supervised learning \and Inversion}

\section{Introduction}
% Seismic prospecting can provide relatively high resolution information of a wide area of the Earth's subsurface. This method has been the primary source of information in oil and gas exploration. However, limitations in acquisition often result in unsatisfactory seismic data with issues like high level of noise, coarsely sampled data, obstacles causing gaps in the data, poor illumination, etc. All of these threats degrade the geometrical features in the seismic data needed to properly images structures in the subsurface. Seismic data processing plays a crucial role in addressing these common issues in the acquired seismic data. Yet, one has to carefully choose and tune available fixed algorithms to address a specific issue found in the data. Yilmaz (2001) showed that errors in seismic data processing could yield erroneous results in imaging, thus may lead to incorrect interpretation of the subsurface.  

Based on the general foundation of algorithms, seismic processing can be classified into two types, namely theory-driven and data-driven approaches. Theory-driven approaches mainly utilize the nature of the travelling seismic waves, exploiting the physics behind them. Numerous methods, which belong to this category, have been proposed, where some of the most notable classic methods include $F$-$x$ domain trace interpolation \cite{spitz1991seismic}, $\tau$-$p$ transform for denoising \cite{turner1990aliasing}, wavelet transform for ground-roll suppression \cite{deighan1997ground}, etc. As pointed out by \cite{hou2021machine}, these "physics-based" algorithms may provide theoretically optimal solutions, though may also face difficulties in dealing with data limitations in acquisition and the presence of phenomena not covered by the physics, like noise. In complex field conditions, admitting poor data, data-driven approaches may come in handy. The direct dependency of the method on data and the freedom to expand beyond the theoretical limitations of physics \cite{sun2020ml} allows for flexibility in handling complicated raw seismic data.

Machine learning (ML) has been a beneficial tool in assisting data-driven seismic processing over the last decade. One of the earliest work in the field was denoising of seismic data using artificial neural network (ANN, \cite{zhang2010seismic}). With the growth of computational power and easier access to data, researchers developed more advanced and better machine learning algorithms geared to image processing and analysis (computer vision) such as generative adversarial network (GAN, \cite{goodfellow2014generative}) and U-Net convolutional neural network \cite{ronneberger2015u}, and, used in audio-related tasks, the Long-Short Term Memory (LSTM) networks \cite{sak2014long}. These networks drove many of the early advances of ML-based seismic processing tasks in the recent years. For example, \cite{ovcharenko2019deep} used a U-Net to extrapolate missing low frequencies to enhance the seismic data spectrum range. \cite{zhang2019automatic} tried to combine "You Only Look Once" (YOLO) and LSTM architecture to automatically pick velocities for normal moveout (NMO) correction. \cite{ovcharenko2020deep} compared U-Net, UGAN (U-Net with adversarial loss), and GMCNN (Generative Multi-column Convolutional Neural Network) in reconstructing seismic data. \cite{fang2021seismic} constrained U-net with a loss from texture segmentation for interpolating seismic data. \cite{borges2021cyclic} used a cyclic learning framework in which CNN networks and cGAN (conditional GAN) are iteratively trained and evaluated on a given seismic processing task to boost its performance. Most recently, \cite{birnie2021potential} adopted Noise2Void (N2V), which is based on CNN architecture to suppress seismic noise in a self-supervised setting. However, most of these studies address a specific seismic processing task, not taking advantage from the fact that most seismic processing tasks utilize similar features and structures embedded in the seismic dataset.

The transformer, from the famous paper \textit{Attention is All You Need} \cite{vaswani2017attention}, is an encoder-decoder self-supervised network architecture, which has proven its robustness and efficiency in language translation tasks. Gaining attention in Natural Language Processing (NLP) community, variants of the Transformer have been developed, e.g. BERT \cite{devlin2018bert}, XLNet \cite{yang2019xlnet}, GPT-3 \cite{brown2020language}, etc., and advances are still being made. The main ingredient of the Transformer is the self-attention blocks, which have been shown in many studies that it could capture, in a detailed manner, global dependencies of the input (e.g. \cite{vaswani2017attention}, \cite{yu2021attention}). Unlike common sequence-to-sequence networks such as LSTM, Transformers do not suffer from computational burden in dealing with long sequences. With these features, its use has been extended to audio and speech representations (e.g. \cite{liu2020mockingjay}, \cite{chi2021audio}). Several studies also showed that Transformer-based architectures are superior to convolutional architectures, making it now renowned and widely used for images in Computer Vision (CV) tasks (e.g. \cite{dosovitskiy2020image}, \cite{he2021masked}). Seismic data could be seen to be similar to audio (time-series) data or image data, because it inherits some of the properties of both. There are several attempts to incorporate self-attention mechanism into the network to aid in capturing global dependencies of seismic data (e.g. \cite{saad2021self}, \cite{yu2021attention}), but only a few applied the complete Transformer-based architecture to seismic data (e.g. \cite{mousavi2020earthquake}), especially in the field of seismic data processing.

In this study, we introduce a new framework for data-driven seismic processing. We, specifically, utilize the Bidirectional Encoder Representation of Transformers (BERT), which was originally developed for NLP tasks, as our baseline network architecture. We call the framework StorSeismic, which highlights the process of "storing the features of seismic data". This framework consists of two steps: pre-training and fine-tuning, as presented in the original paper. The network is pre-trained on a seismic dataset in a self-supervised manner, allowing the network to capture and store key features (i.e. geometrical) of seismic data. Afterwards, we could fine-tune the pre-trained model for specific downstream tasks of seismic processing such as trace interpolation, denoising, first-break picking, etc. Through this framework, we could obtain a task-specific network for a given seismic processing task more efficiently and robustly than existing approaches. We also demonstrate here how to utilize such a framework on field data, which is often a big challenge for ML considering that labels are expensive, even sometimes impossible to obtain. Thanks to the integrated framework of StorSeismic, the requirements of the field labels in a common supervised training could be accommodated by the prior knowledge that the model acquires by including the field data in the pre-training step of the framework, which is performed in a self-supervised setting. 

The contribution of this study includes the following:
\begin{itemize}
    \item We propose a new network architecture, based on BERT, for seismic processing and a framework of utilizing it in pre-training and fine-tuning steps.
    \item We show the scalability of a single pre-trained model over a wide range of seismic processing tasks, with a minimum change to the model's parameter.
    \item We show its ability to adapt to label-less field data by fine-tuning on labeled synthetic data.
\end{itemize}

The paper is structured as follows. In Section \ref{sec:theory}, we detail the proposed network architecture and the trace reconstruction procedure in the pre-training as the foundation of the framework. Testing of our proposed network and framework along with some analysis will be presented in Section \ref{sec:test}. Section \ref{sec:data} will include the details of the dataset that we use and the workflow we apply to address the label-less field data. In Section \ref{sec:examples}, we show results from the application of this method on synthetic and field data. We discuss implementation details, specifically with respect to fine-tuning, as well as some of the method's limitations in Section \ref{sec:discussion}. Finally, we summarize our work in Section \ref{sec:conclusion}.

\section{Theory}
\label{sec:theory}
In this section, we first detail the architecture of StorSeismic and then explain how we will use it with seismic data. We will then discuss the pre-training procedure, which is key to storing the seismic data features.

\subsection{Network architecture}
\begin{figure}[h]
    \centering
    \includegraphics[width=8.5cm]{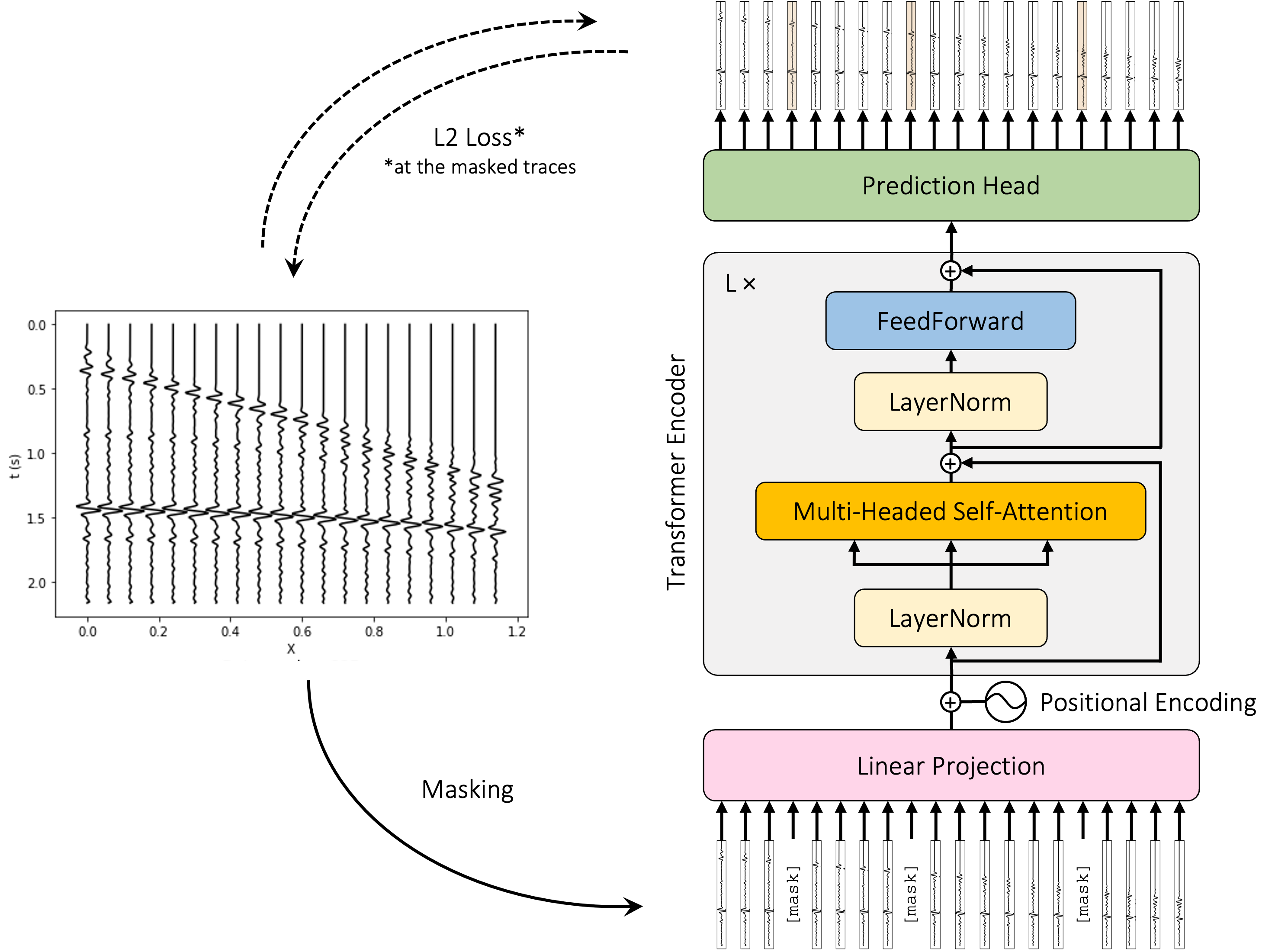}
    \caption{The architecture of StorSeismic. Masked seismic data are linearly projected and supported with positional encoding before passing the output through a series of attention blocks. At the prediction head, the traces are reconstructed, then we measure the loss only on the masked traces.}
    \label{fig:fig1}
\end{figure}
An overview diagram of the StorSeismic architecture is presented in Figure \ref{fig:fig1}. Here, we follow the design of the encoder in the Transformer architecture \cite{vaswani2017attention} and BERT \cite{devlin2018bert} with some modifications to accommodate the input of seismic data, considering that the original architecture catered to inputs given by sequences of words. 

The model is composed of three main parts: the \textit{embedding block}, \textit{encoder block(s)}, and \textit{prediction head}. Suppose we have 2D seismic data $d\;\epsilon\;\mathbb{R}^{\mathrm{X} \times \mathrm{\Tau}}$ as our input, where $\mathrm{X}$ is the number of traces (channels) and $\Tau$ is the number of time samples. We mask a portion of the traces in the seismic data to simulate missing data, for which the reason behind this step will be explained in the following section. We denote the masked seismic data as $d_m$. Instead of using word embedding as in the original BERT and Transformer models, we use a linear layer $E_1$ to project $\Tau$ to the hidden dimension H. This allows the network to find a better representation of the data in a latent space.

Unlike LSTMs, Transformers do not have "memory cells" to recognize the position of the sequences. To overcome this issue, a "positional encoding" $E_2$ is combined with the linearly projected data. We adopt the sinusoidal positional encoding function used with the original Transformer \cite{vaswani2017attention}:

\begin{equation}
    \label{eq:eq1}
    E_2(pos, i) =
    \begin{cases}
        \sin(pos/10000^{2i/H}), & \text{if $i$ is even} \\
        \cos(pos/10000^{2i/H}), & \text{if $i$ is odd}
    \end{cases}
\end{equation}

where $pos$ is the relative position of the traces (i.e. offset index). Therefore, the output of the embedding block is:

\begin{equation}
    \label{eq:eq2}
  \hat{E} = E_2(pos, E_1(d_m)).  
\end{equation}

For the next step, the embedded traces $\hat{E}$ are passed through a series of L encoder blocks. Each of these blocks contains a multi-headed self-attention (MSA) layer, a position-wise feed-forward (FFN) layer, and layer normalization along with residual connections, like in Figure \ref{fig:fig1}. Inside the MSA layer, the normalized embedded traces are first linearly projected into three different matrices namely query ($Q$), key ($K$), and value ($V$):

\begin{equation}
    \label{eq:eq3}
    Q = W_Q\hat{E},\;\;\; K = W_K\hat{E},\;\;\;V = W_V\hat{E},
\end{equation}

where $W_Q$, $W_K$, and $W_V$ are learned weights of the linear projection for the query, key, and value matrices, respectively. Then, the attention operation is defined as:

\begin{equation}
    \label{eq:eq4}
    A_t = Attention(Q, K, V) = softmax\left(\frac{QK^T}{\sqrt{H}}\right)V.
\end{equation}

 The softmax operation can be thought of as a weighting matrix applied to $V$, and the matrix multiplication of $Q$ and $K$ is a measure of similarity (i.e. cross-correlation) between the traces (or their key features). Thus, each output (i.e. trace) in the sequence is a weighted summation of the input sequences (traces or their features). Note that $A_t$ still has the same first dimension as the input seismic channels $d$, i.e. $A_t\;\epsilon\;\mathbb{R}^{\mathrm{X} \times \mathrm{H}}$. These operations are distributed over the number of attention heads A, which are done by splitting the input beforehand, and concatenating them back afterwards.

Matrix $A_t$, after adding a residual connection and normalizing it using a layer normalization, acts as the input of the FFN layer. The feed-forward network (FFN) is a simple two-layered neural network with a GELU activation function in between. The output of the FFN layer is denoted as:

\begin{equation}
    \label{eq:eq5}
    F = FFN(A).
\end{equation}

Once again, a residual connection is applied to $F$. Finally, the output of the final encoder block goes into the prediction head $P$. The role of this prediction head is to project the hidden dimension H back to the input dimension, $\Tau$. Thus, we could make use of a simple linear layer. The final output $\hat{d}$ can be written as:

\begin{equation}
    \label{eq:eq6}
    \hat{d} = P(F),
\end{equation}

which is given by the reconstructed seismic data in the pre-training stage. This prediction head is detachable and used only during pre-training stage. For the fine-tuning stage, the prediction head is replaced with NN layer(s), needed for a given downstream task.
 
\subsection{Trace reconstruction in the pre-training stage}

As \cite{rogers2020primer} highlighted, a pre-training followed by fine-tuning workflow constitute the components of BERT, where the former supposed to equip BERT with task-independent knowledge on, for example, sentences. Similarly, for seismic data, we expect StorSeismic to capture the features of seismic data during the pre-training step. BERT was initially pre-trained on two tasks, namely Masked Language Modeling (MLM) and Next Sentence Prediction (NSP), simultaneously. In this study, we borrow the concept of MLM to pre-train our model.

We treat the seismic section (i.e shot gather) as a sentence and the traces as individual words. Thus, MLM training in NLP is equivalent to reconstructing masked traces in a seismic data. Let $d$ be the observed seismic data and $d_m$ is the masked version of the data. So:

\begin{equation}
    \label{eq:eq7}
    d_m = Md,
\end{equation}

where $M$ is the masking operator. The objective here is to formulate a neural network operator (BERT) that converts the masked seismic data $d_m$ back to $d$, where in our machine learning setting $d_m$ and $d$ correspond to the input and the label, respectively. We propose our model, StorSeismic, in the pre-training stage as a reconstruction operator denoted by $SS$. The output of this operator, given masked seismic data $d_m$ as input, is the predicted original data $\hat{d}$:

\begin{equation}
    \label{eq:eq8}
    \hat{d} = SS(d_m).
\end{equation}

During pre-training, we teach our model to reconstruct the missing traces. Hence, we use a selection operator $M' = I - M$, where $I$ is the identity matrix, to only select the location of the masked traces for the loss function. Using an L2 norm, we train the network by minimizing:

\begin{equation}
    \label{eq:eq9}
    \ell_{\theta} = ||M'\hat{d_{\theta}} - M'd||_2,
\end{equation}

where $\theta$ is the neural network parameters.

\section{Test and Analysis}
\label{sec:test}

We first test the approach on simple benchmarking synthetic data, SNIST \footnote{\url{https://github.com/LukasMosser/SNIST}}, modeled from 1D velocity models made up of random layers. The data contain 600 and 150 shot gathers for training and testing, respectively, with their corresponding velocity profiles. Each shot gather contains 20 traces (channels) sampled at 8 ms time interval for 271 time steps. We scale the amplitudes of the shot gathers to (-1, 1).

We initialize our model StorSeismic with L = 4 (attention layers), H = 256 (hidden layer dimension), and A = 4 (attention heads), resulting in a total of 3.3M trainable parameters. This configuration corresponds to BERT\textsubscript{MINI} model in the original implementation of BERT \footnote{\url{https://github.com/google-research/bert}}. We will show two fine-tuning tasks on the SNIST dataset (denoising and velocity estimation). The code to build the StorSeismic model and to reproduce this experiment will be made available on our Github repository \footnote{\url{https://github.com/swag-kaust/storseismic}}.

\subsection{Pre-training}
We augment the SNIST dataset by slightly shifting the time samples, changing the polarity, and masking the traces at different positions of the input, which expands the dataset to 36,000 and 9,000 samples for training and testing, respectively. The masking proportion is 15\% in each sample (3 out of 20 traces), and we follow the masking procedure in BERT \cite{devlin2018bert} in which 80\% of the masked region is filled with a [mask] token (i.e. random numbers drawn from a Gaussian distribution), 10\% were replaced with values from a trace at a different position, and 10\% were kept the same. A batch size of 256 was used in the pre-training, and we optimize the model parameters for 400 epochs using RAdam optimizer \cite{liu2019variance} with a learning rate of 5e-4, which took approximately 1 hour on a Quadro RTX 6000 GPU. 

\begin{figure}[!h]
    \hspace{1.3cm}
    \subfloat[]{\includegraphics[width=11cm]{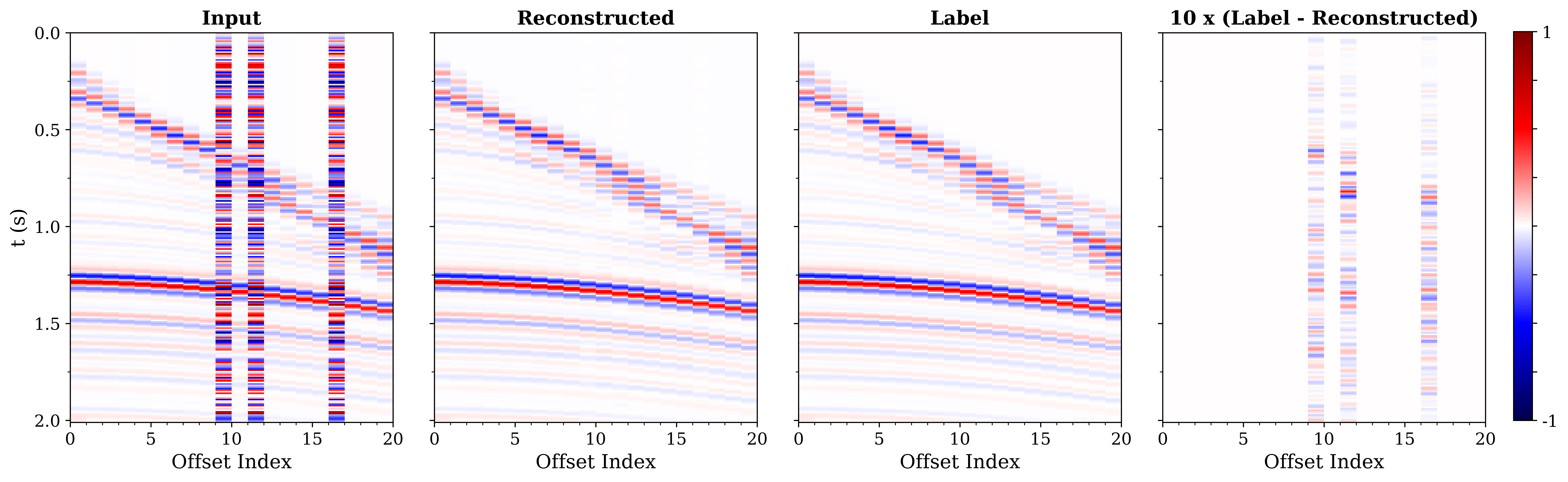} \label{fig:fig2a}}
    \newline
    \centering
    \subfloat[]{\includegraphics[width=11cm]{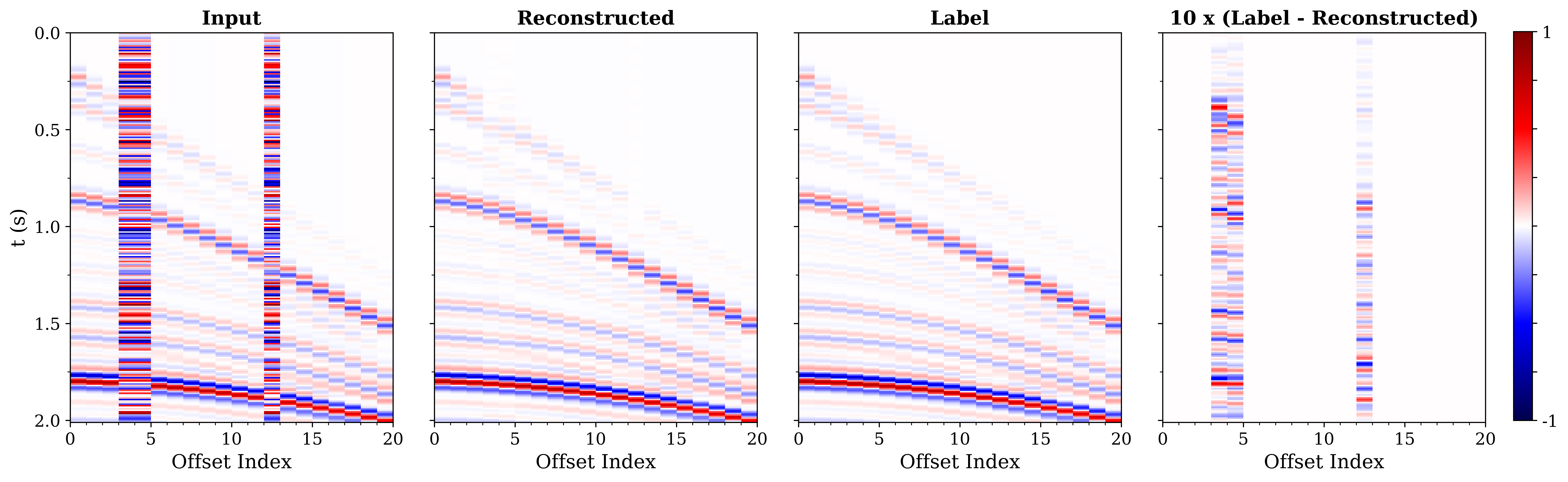} \label{fig:fig2b}}
    \caption{Two examples of reconstruction of masked seismic shot gathers from the testing set (a and b).}
    \label{fig:fig2}
\end{figure}

Examples of the prediction in the test set is shown in Figure \ref{fig:fig2} (a and b). We achieve a Mean Squared Error (MSE) of 8e-5 in the test set (at the masked traces). Although we could utilize the pre-trained model to perform a trace interpolation task, the main reason behind this pre-training is to allow StorSeismic to learn the features of the data, specifically on how the traces are related to each other, and store such information to be used in the subsequent fine-tuning task.

\subsection{Denoising}
An important task in seismic processing workflows is noise removal (i.e. denoising). For this analysis we only add Gaussian noise (40\% were added with 1-sigma of noise and 40\% were added with 2-sigma of noise), and we use only polarity reversal for augmentation, resulting in 1,200 training samples and 300 testing samples (much lower than in the pre-training). With respect to the architecture of $SS$, we only replace the prediction head with a zero-initialized linear layer of the same size in the pre-trained model, and use a loss function for fine-tuning given by an L2 norm, now, over the whole shot gathers. With a batch size of 16, a RAdam optimizer, and a learning rate of 5e-4, the model required 65 epochs and 4 minutes to fine-tune, compared to one hour of pre-training.

\begin{figure}[h]
    \centering
    \subfloat[]{\includegraphics[width=0.5\textwidth]{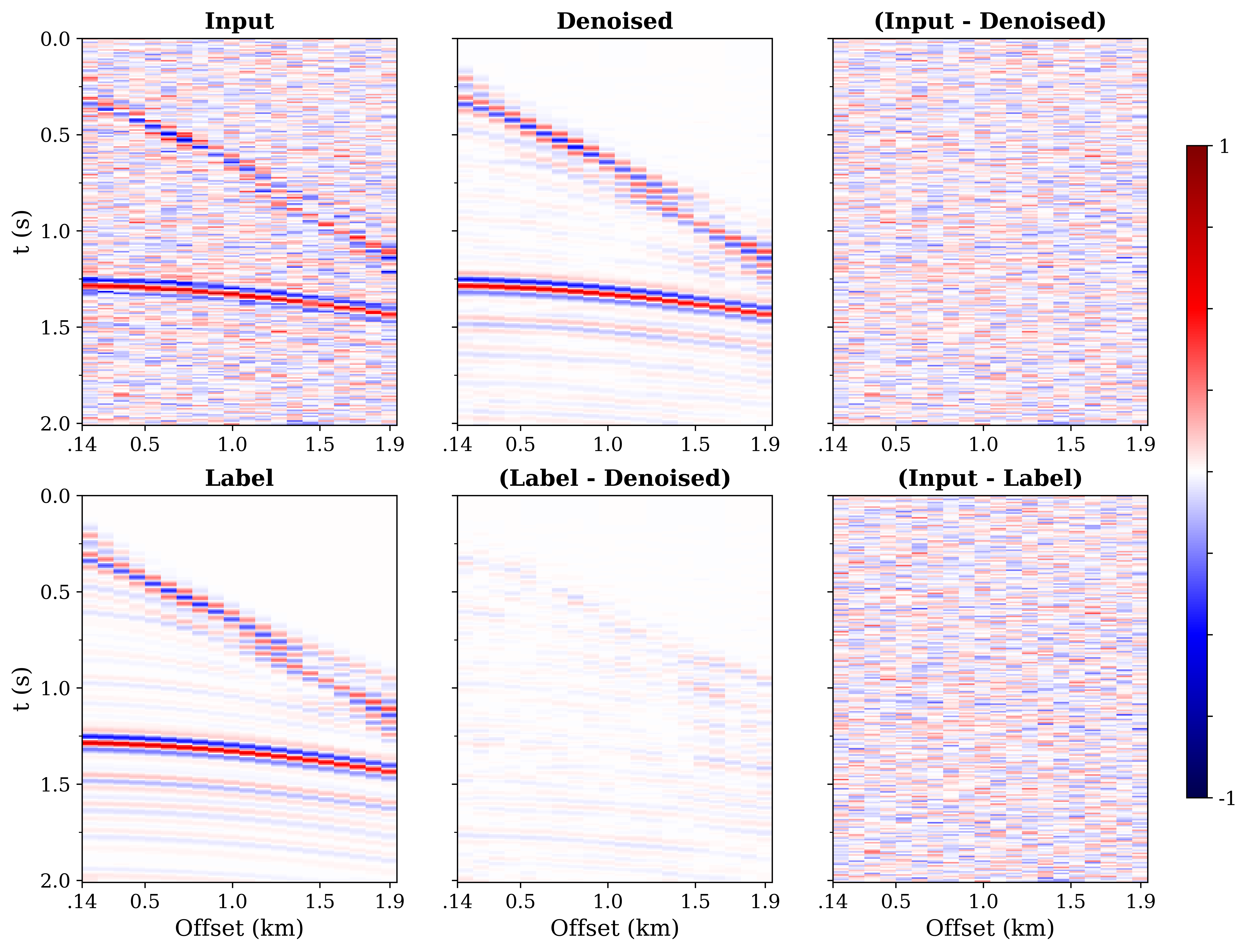} \label{fig:fig3a}}
    \subfloat[]{\includegraphics[width=0.5\textwidth]{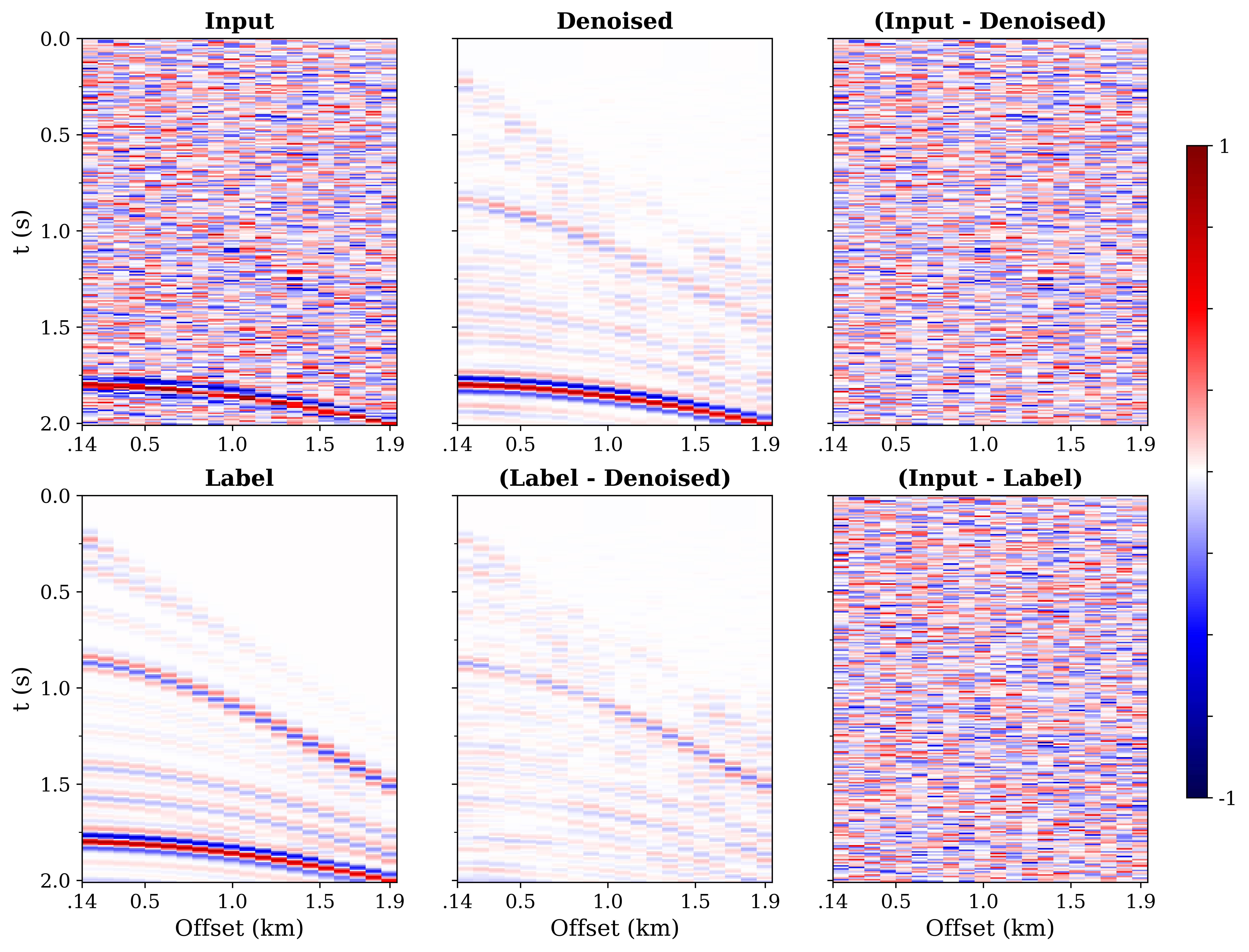} \label{fig:fig3b}}
    \caption{a) Examples of denoising with 1-sigma noise (corresponds to the test shot gather in Figure \ref{fig:fig2a}) from the fine-tuned model. b) Another example, but on data with 2-sigma noise (corresponds to the test shot gather in Figure \ref{fig:fig2b}).}
    \label{fig:fig3}
\end{figure}

The denoising performance of the model for 1-sigma and 2-sigma of noise are shown in Figure \ref{fig:fig3a} and Figure \ref{fig:fig3b}, respectively. We obtain an MSE of 7e-4 on the test set, compared to an MSE of 1.4e-3 prior to fine-tuning, which demonstrates the fine-tuning ability to adapt to a task. For 1-sigma of noise, the noise was reasonably removed from the data with a little signal leakage. 

\subsection{Velocity estimation}
For the second downstream task, we will try to estimate the 1D velocity profiles from a shot gather as the input, with both inputs and labels are available in the SNIST dataset. The same training set up as the denoising task is used, except for the loss function, which for this task, we used the L1 norm between the predicted (taken from the first output in the sequence, i.e., the zero offset) and the true velocities. The training (fine-tuning) took 51 epochs and 3 minutes to converge.

\begin{figure}[h]
    \centering
    \subfloat[]{\includegraphics[height=6cm]{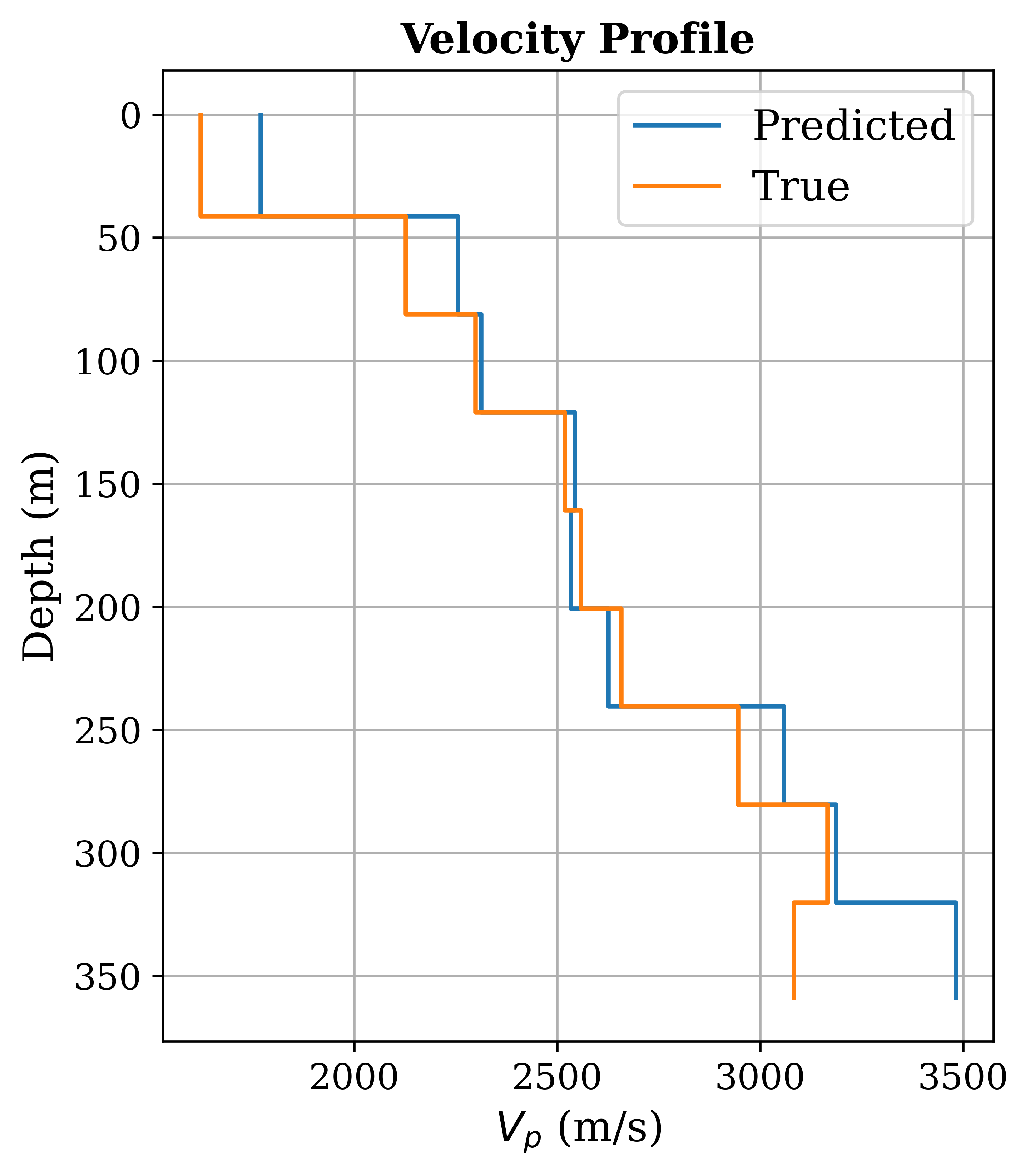} \label{fig:fig4a}}
    \subfloat[]{\includegraphics[height=6cm]{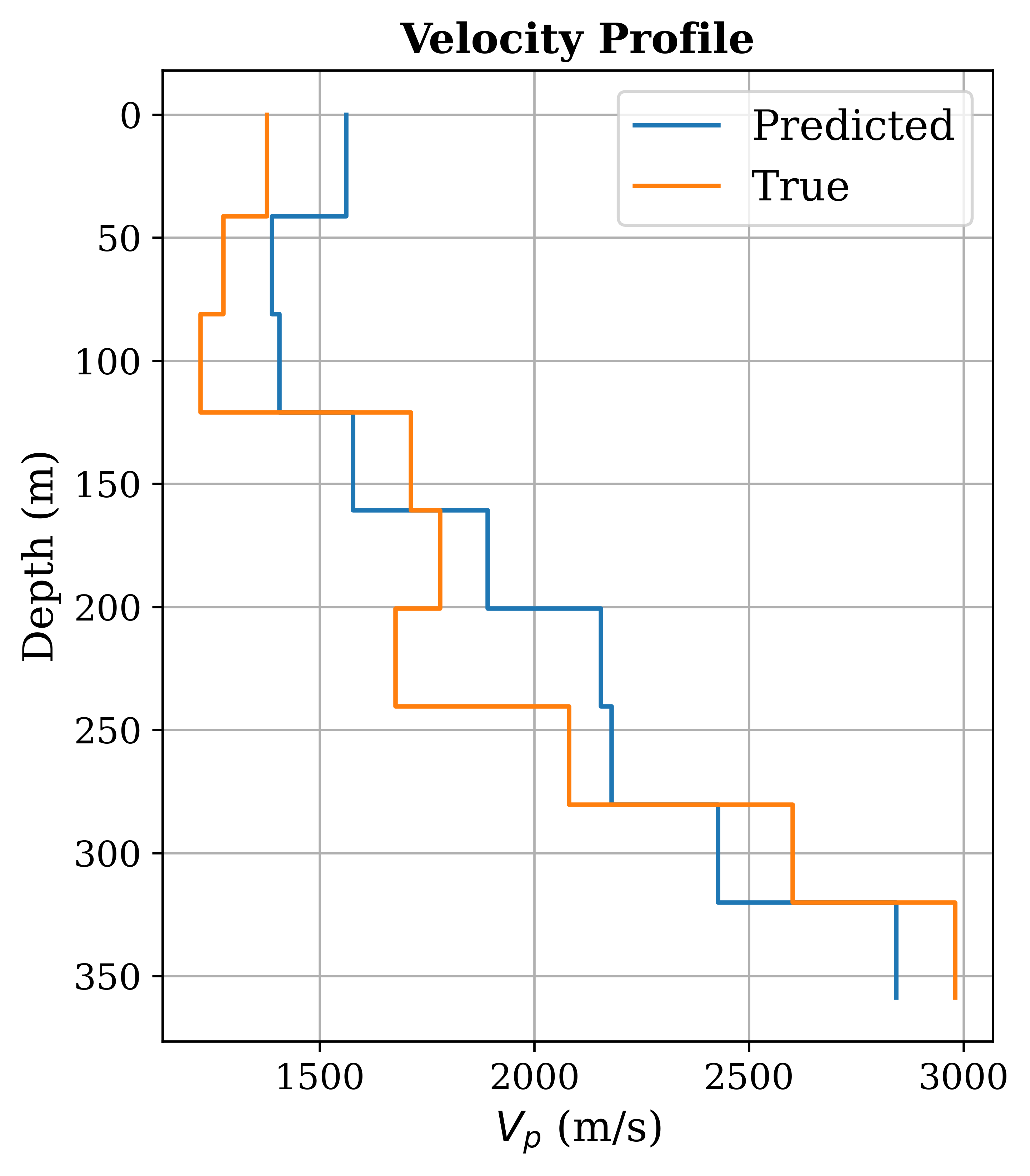} \label{fig:fig4b}}
    \caption{Two examples of velocity estimation on SNIST data. The input correspond to shot gathers shown in Figure \ref{fig:fig2}. Blue line represents the predicted velocity and orange line represents the true velocity.}
    \label{fig:fig4}
\end{figure}

Figure \ref{fig:fig4} shows two examples of the result on the test data. Overall, the model reached a Mean Absolute Error (MAE) of 112 m/s in the test set, and we observed that the model is able to estimate the values of the velocities reasonably well, especially the trend.

\subsection{Analysis}
\label{sec:analysis}
\begin{figure}[h]
    \centering
    \subfloat[]{\includegraphics[width=5cm]{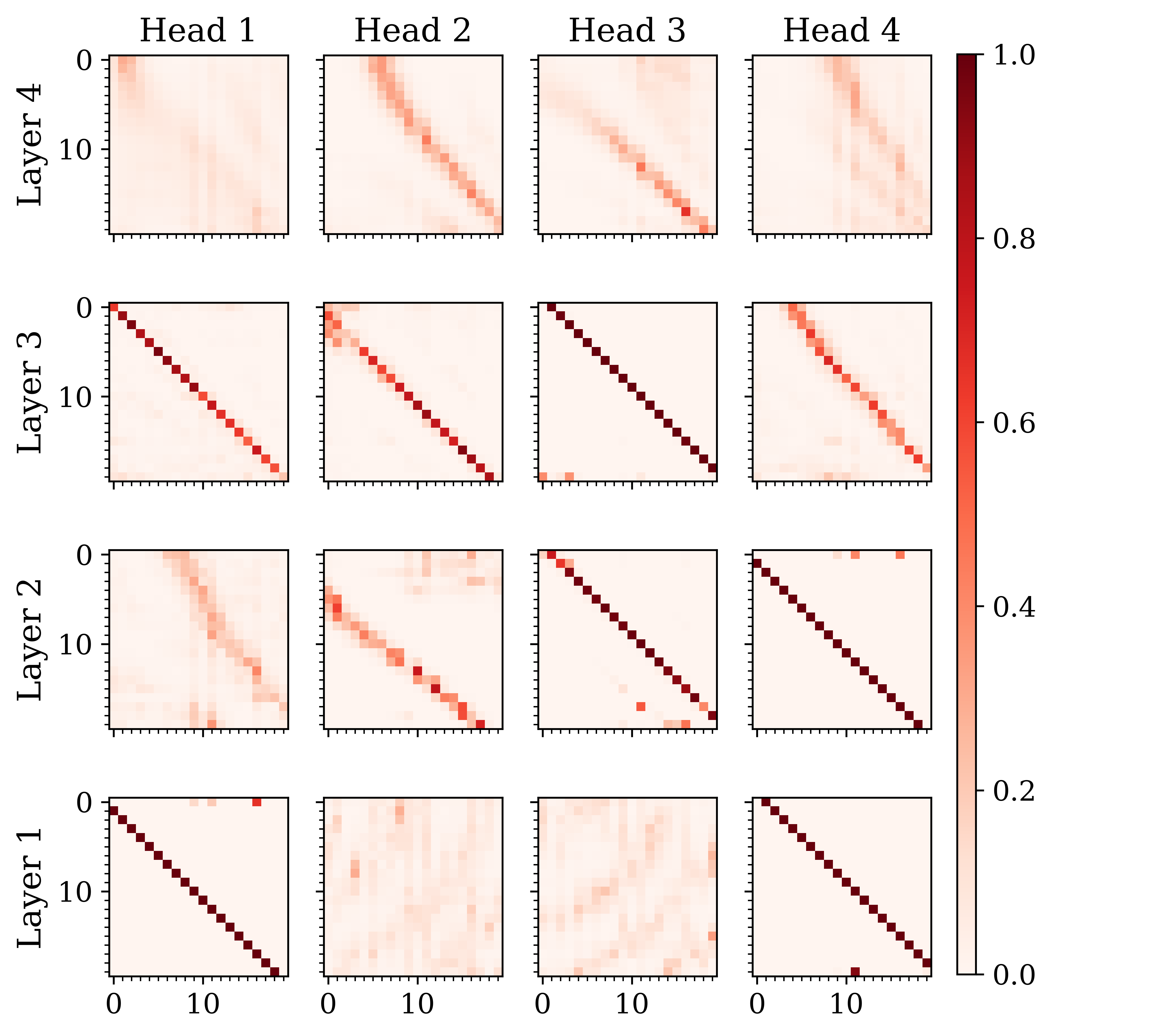} \label{fig:fig5a}}
    \subfloat[]{\includegraphics[width=5cm]{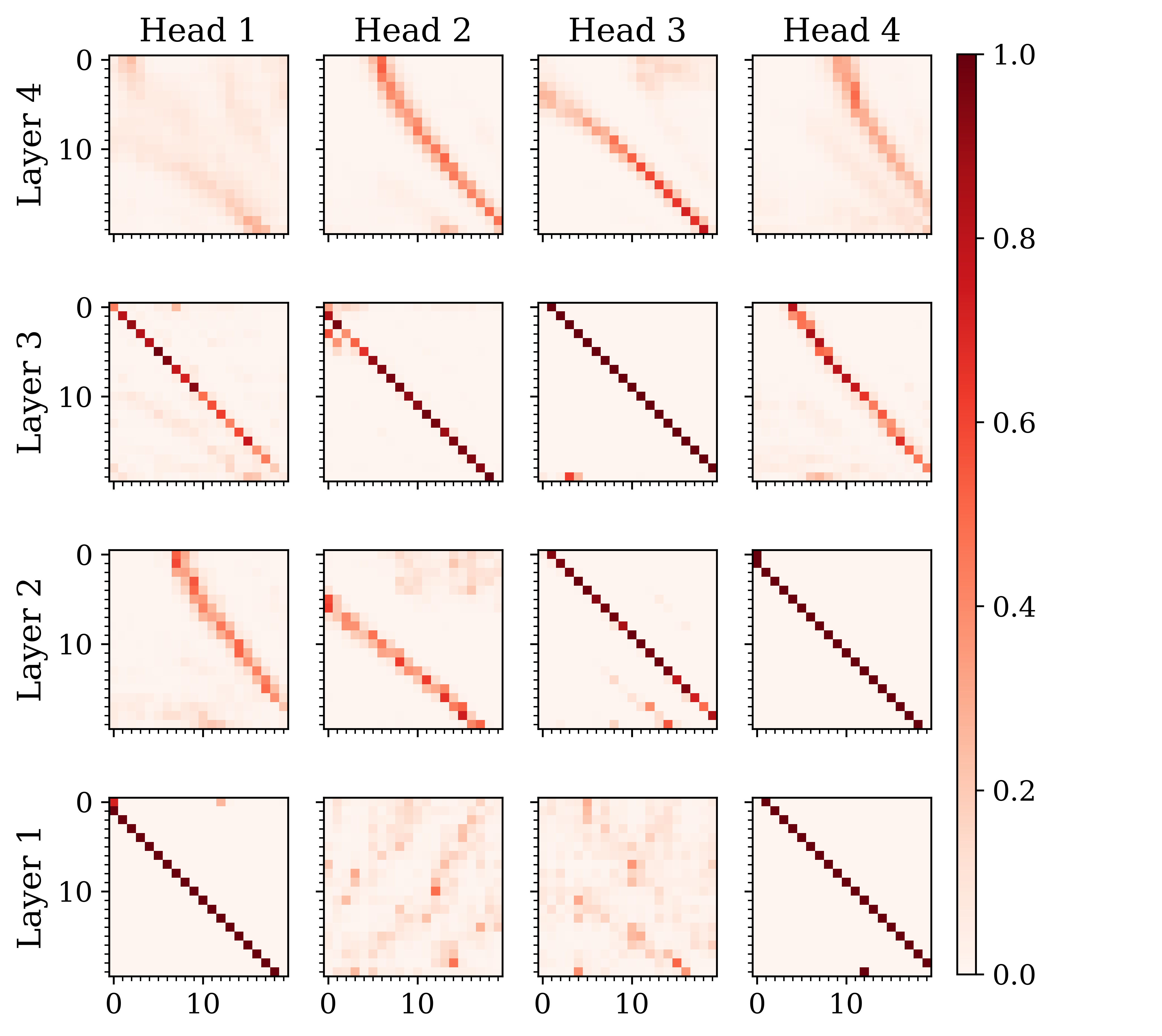} \label{fig:fig5b}}
    \subfloat[]{\includegraphics[width=5cm]{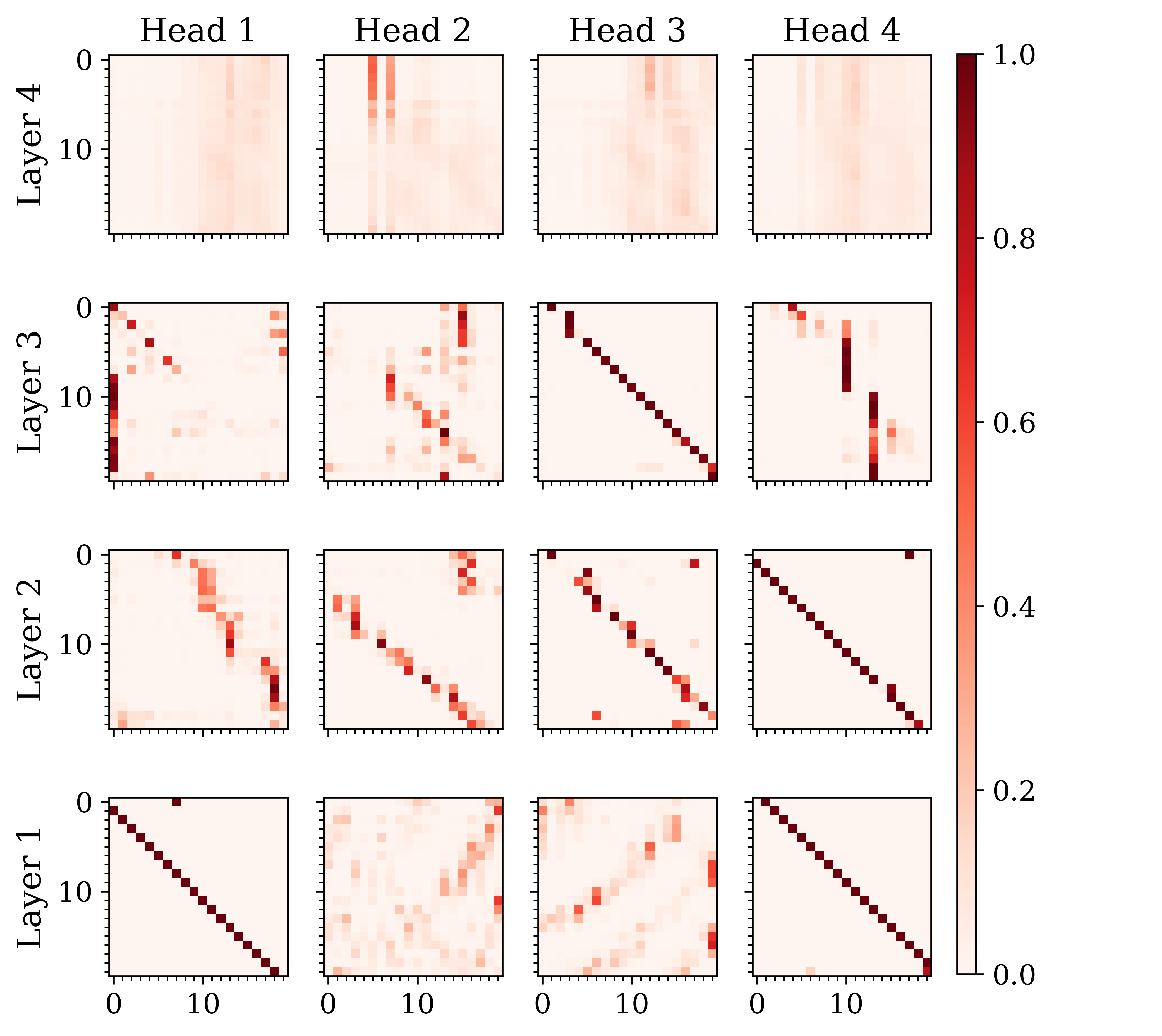} \label{fig:fig5c}}
    \caption{The resulting attention maps of: a). SNIST pre-training (corresponds to Figure \ref{fig:fig2a}), b). SNIST denoising fine-tuning (corresponds to Figure \ref{fig:fig3a}), and c). SNIST velocity estimation fine-tuning (corresponds to Figure \ref{fig:fig4a}).}
    \label{fig:fig5}
\end{figure}

We will now try to understand why did this framework adapt fast to various tasks. Since the only opportunity for the various traces to communicate is the attention layers, we will focus on them. Attention maps, which consist of the weights of the attention heads (softmax operation on Equation \ref{eq:eq4}), are provided in Figures \ref{fig:fig5a}, \ref{fig:fig5b}, and \ref{fig:fig5c} for the corresponding examples shown in Figure \ref{fig:fig2a}, \ref{fig:fig3a}, and \ref{fig:fig4a}, respectively. As we can see from the plots, the pattern of the attention maps changes accordingly to the given task, especially in the deeper layers. Meanwhile, we observe similarities of the attention maps in the shallower layers after the fine-tuning training (Figures \ref{fig:fig5b} and \ref{fig:fig5c}) with that of the pre-training (Figure \ref{fig:fig5a}). This shows that while the pre-trained model adapts to a specific task, it still inherits the information learned at the pre-training step. 

\begin{figure}[h]
    \centering
    \includegraphics[width=8.5cm]{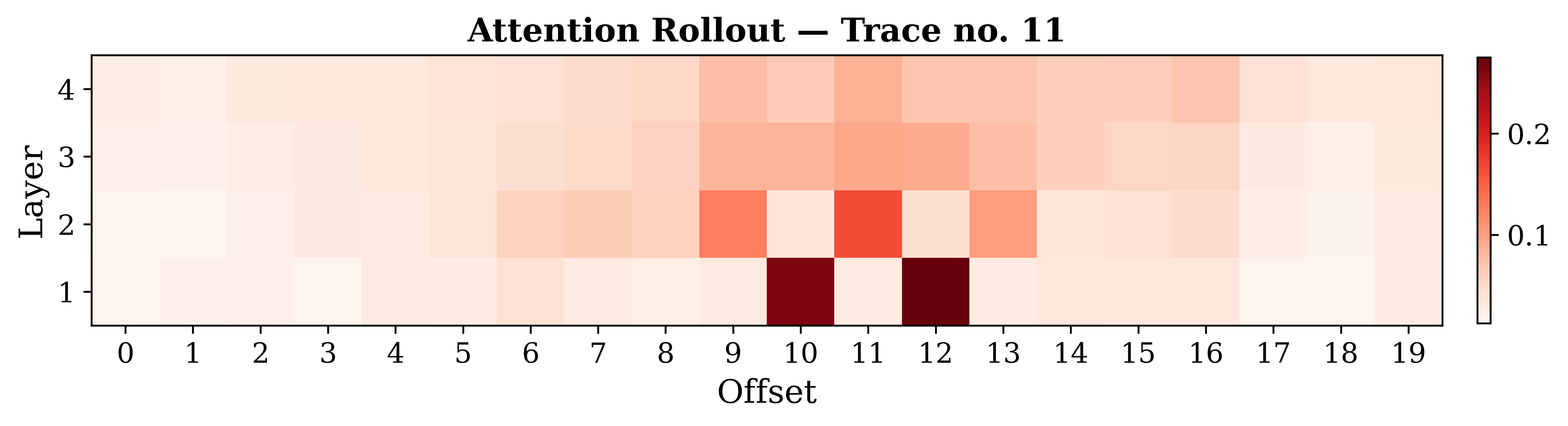}
    \caption{Attention Rollout (top) for trace no. 11 calculated from the attention maps in Figure \ref{fig:fig5a}.}
    \label{fig:fig6}
\end{figure}

One way to interpret the attention maps is through the Attention Rollout \cite{abnar2020quantifying}, which in short is a recursive matrix multiplication of the average of the attention maps (we use attention maps in Figure \ref{fig:fig5a} as an example) through each layer (Figure \ref{fig:fig6}). From the plot, we can clearly see that the network learns to get information for the masked trace at offset index 11 primarily from the adjacent traces at shallow attention layers, then from other traces at deep layers of the network. We also observe this pattern on other traces. This demonstrates the ability of the network to capture both global and local features in the seismic data.

\section{A field data framework}
\label{sec:data}
\begin{figure}[!h]
    \centering
    \includegraphics[width=13cm]{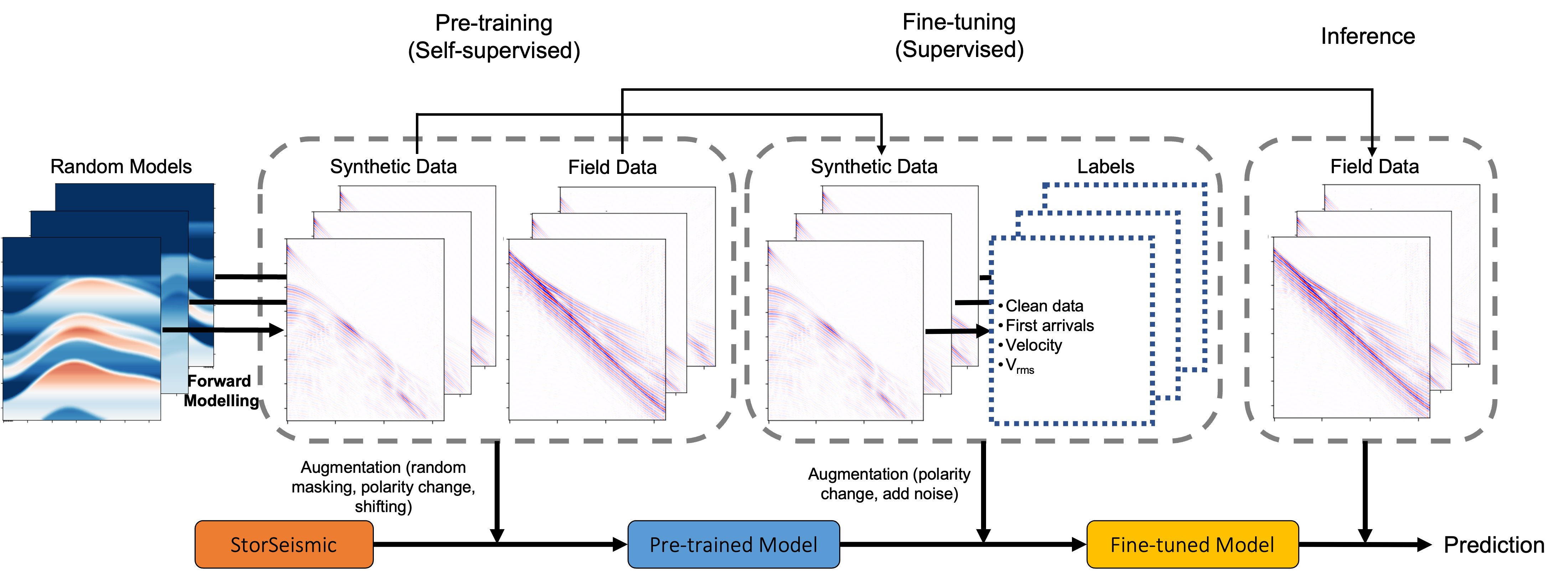}
    \caption{The field dataset StorSeismic workflow consisting of three steps: pre-training in a self-supervised setting using a mix of synthetic and field data, fine-tuning in a supervised setting using synthetic data and their corresponding labels, and finally inference on field data for the task it has been fine-tuned for.}
    \label{fig:fig7}
\end{figure}

The proposed framework allows for easy adaptation to field data, whether the field data include labels or not. For label-free field data, we employ the following strategy. First, we generate synthetic data using random subsurface models, which are tailored to represent, as much as possible, the Earth properties influencing our field data. Figure \ref{fig:fig7} shows examples of the random models, and detailed description of how they have been generated can be found in \cite{ovcharenko2021data}. The distribution of these models can be guided by our prior knowledge of distribution of velocity models representing the area, and in particular the area covered by the seismic data or guided by well information. In the pre-training stage, we use a mix of synthetic and field data in a self-supervised manner, with an intention to teach the model to store the features of both the synthetic and the field data, as if they were one dataset, which is inspired by \cite{nowruzi2019much}. Considering that for synthetic data we often model the input features from their labels (like adding noise, or waveform simulation), we use them to fine-tune the pre-trained model in a supervised manner to perform a specific task. In addition, to preserve the imprint of the field data that the model was exposed to during the pre-training, we only fine-tune the last two layers of the pre-trained model (see Section \ref{sec:technique}). Finally, we use the fine-tuned model to make predictions on the field dataset. This workflow is illustrated in Figure \ref{fig:fig7}.

\section{Application}
\label{sec:examples}

The field data contain 1,824 shot gathers from a marine seismic acquisition which contains 324 traces (channels) sampled at 16 ms time interval for 376 time steps each shot. To balance the proportion of field and synthetic data in the pre-training, we generate the same number of shot gathers (i.e. 1,824 samples) using 1,024 random subsurface models (a sample is shown in Figure \ref{fig:fig7}) with an elastic forward modeling software \cite{kohn2011time}. We will use the high frequency data (obtained by applying a high-pass filter with a corner frequency of 4 Hz), which we refer to as "SHF" and "FHF" for synthetic and field data, respectively, as inputs for all the training. The amplitude in all sets are normalized to (-1, 1).
We will perform the two tasks shown previously (denoising and velocity estimation) and extend two more tasks (first-break picking and NMO).

\subsection{Pre-training}
\label{sec:pretrain_field}
Following the workflow depicted in Figure \ref{fig:fig7}, we split each of SHF (synthetic set) and FHF (field set) into 1,276 and 548 samples for training and testing, respectively, and we use an equal mix of SHF and FHF as the dataset for the pre-training. Several augmentation procedures are applied to the dataset, which include slightly shifting the time samples, reversing the polarity, and masking the traces at different positions of the shot gather, resulting in expanded data of 45,936 and 19,728 samples for training and testing, respectively. We randomly mask 48 out of 324 (15\%) traces in each shot gather. We train the model with a batch size of 128 and using the RAdam optimizer with a learning rate of 5e-4. Convergence is reached after 249 epochs, which took 12.7 hour on a Quadro RTX 8000 GPU.

\begin{figure}[h]
    \hspace{1.3cm}
    \subfloat[]{\includegraphics[width=11cm]{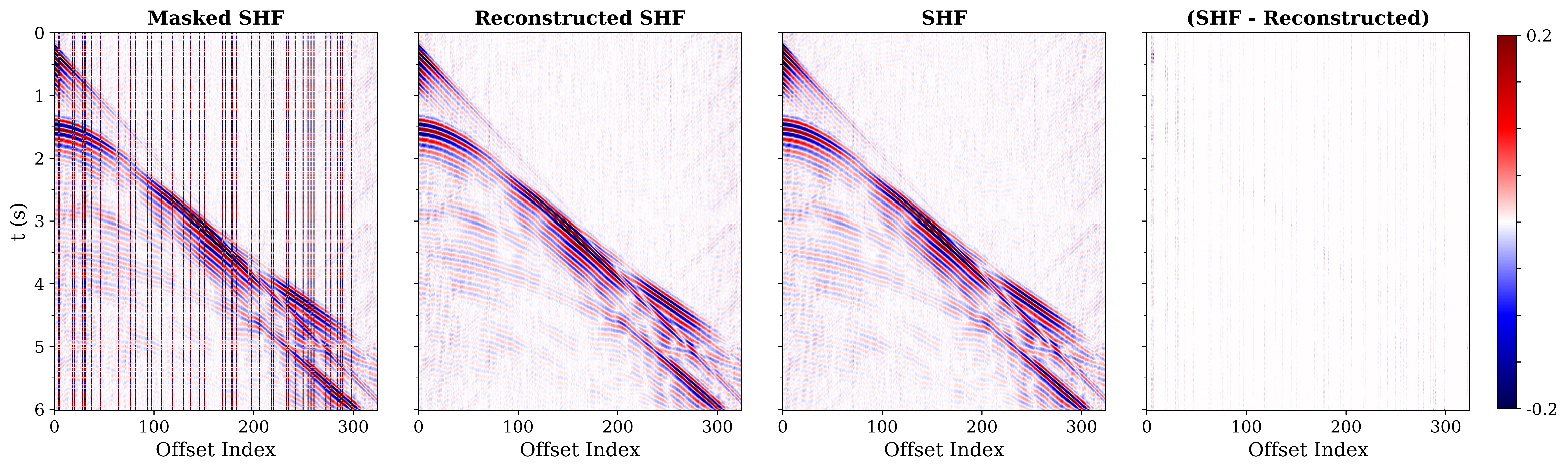} \label{fig:fig8a}}
    \newline
    \centering
    \subfloat[]{\includegraphics[width=11cm]{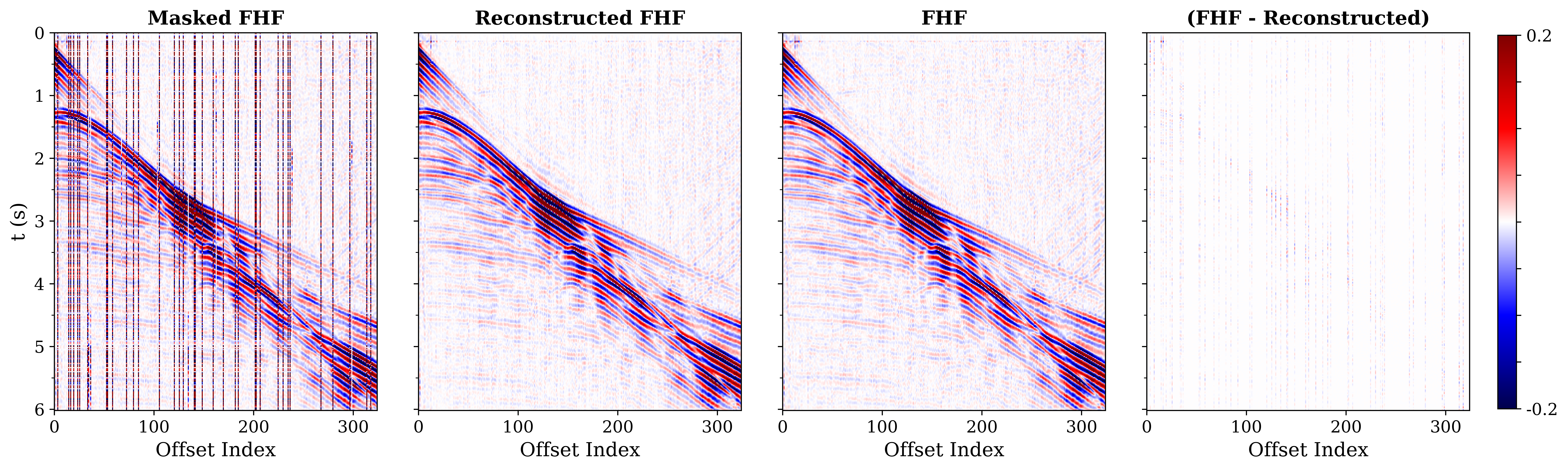} \label{fig:fig8b}}
    \caption{Reconstruction of masked seismic shot gathers from the testing set for synthetic data (SHF, a) and field data (FHF, b) examples.}
    \label{fig:fig8}
\end{figure}

Pre-training results on synthetic and field data test samples are shown in Figure \ref{fig:fig8}. The model reached a Mean Squared Error (MSE) of 1.7e-5 in the test set (at the masked traces). Again, the ability of the model to reconstruct the masked traces demonstrates that the NN model is able to recognize the features of the data, and it is ready to use them in the subsequent fine-tuning tasks.

\subsection{Denoising}
\label{sec:denoising_field}
We prepare the training set for this fine-tuning task by reusing the original partitioned SHF (synthetic dataset) and augment it with a polarity reversal, which expands the original dataset into 2,552 and 1,096 samples for training and testing, respectively. Then, we randomly sample noise from FHF and add them to the SHF samples, to represent the input to the supervised denoising task. The noise free data act as labels. We replace the prediction head of the pre-trained model with a randomly-initialized linear layer, then we fine-tune the model using a batch size of 16, RAdam optimizer with a learning rate of 5e-4, and an MSE loss between the output (i.e. denoised data) and the clean data. The fine-tuning took 900 epochs and 3 hours to train, compared to 10 hours of pre-training.

\begin{figure}[h]
    \centering
    \subfloat[]{\includegraphics[width=0.5\textwidth]{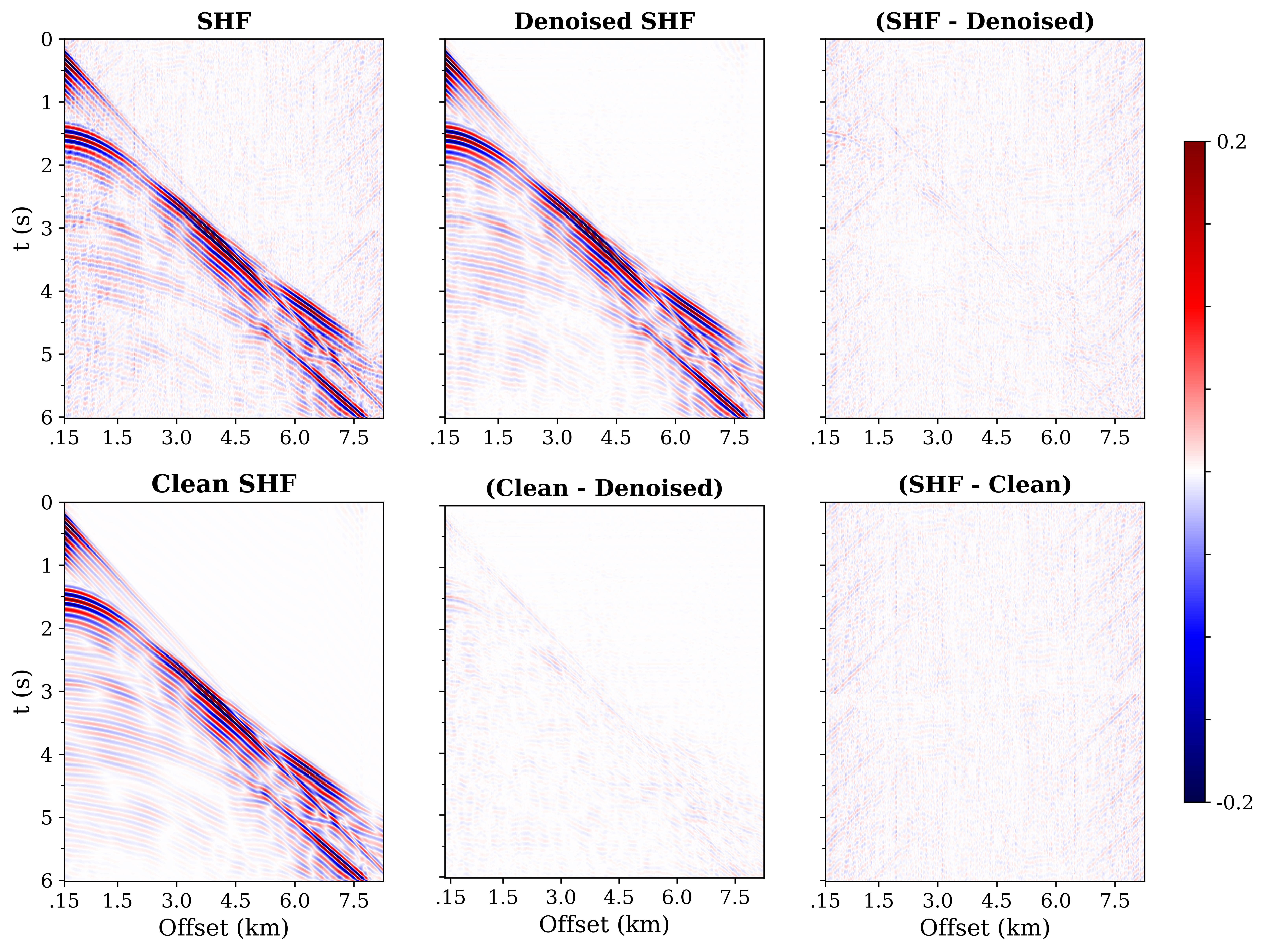} \label{fig:fig9a}}
    \subfloat[]{\raisebox{0.5\height}{\includegraphics[width=0.5\textwidth]{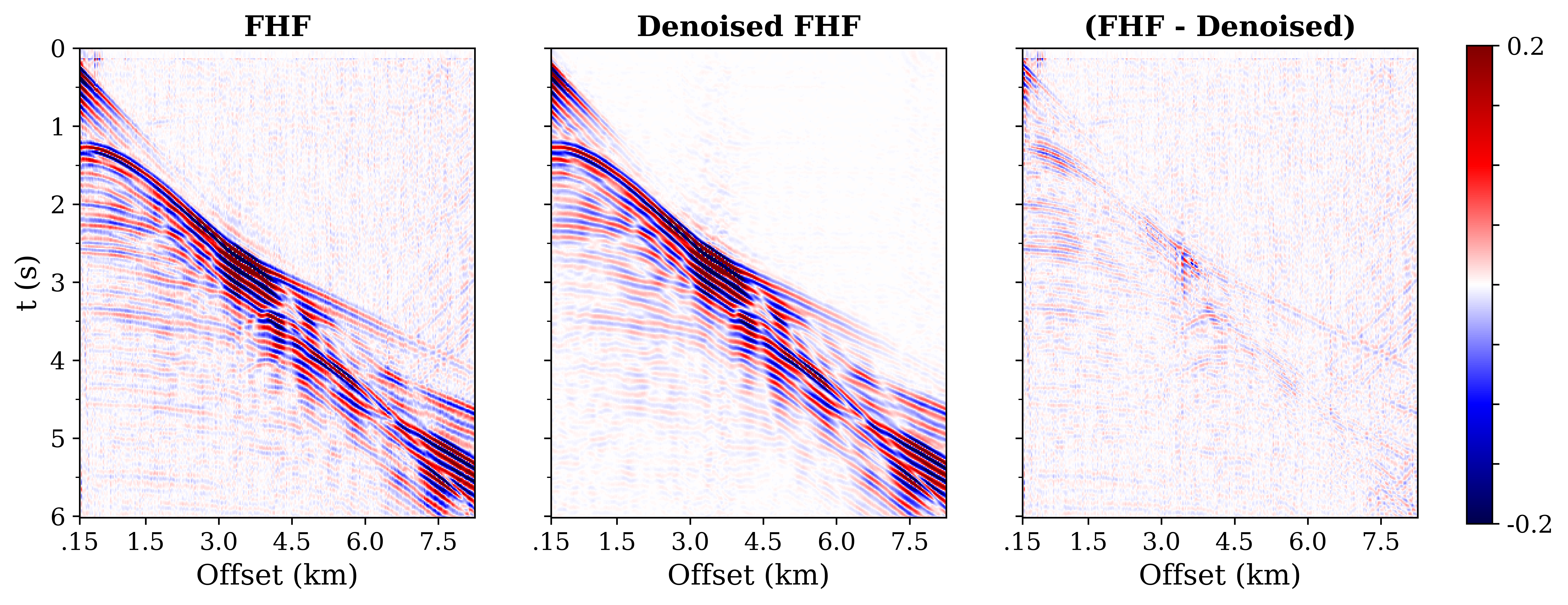}} \label{fig:fig9b}}
    \caption{Examples of denoising using the fine-tuned model on synthetic data (SHF, a) and field data (FHF, b). The input correspond to SHF and FHF shown in Figure \ref{fig:fig8}.}
    \label{fig:fig9}
\end{figure}

We achieved a test MSE of 2e-5 with examples of the result of denoising displayed in Figure \ref{fig:fig9}. The model managed to remove the added noise (Figure \ref{fig:fig9a}, top right and Figure \ref{fig:fig9a}, bottom right) with minimum signal leakage (Figure \ref{fig:fig9a}, bottom mid). The performance on the FHF is presented in Figure \ref{fig:fig9b}, demonstrating the model's ability to remove the noise, especially the back-scattered noise, while preserving the signal.

\subsection{Velocity estimation}
\label{sec:velpred_field}
In this task, we will try to predict a vertical velocity for each shot gather. To do so, we average the true velocity laterally from the shot position to half the maximum offset (which we will refer as "mean velocity" for simplicity). We use a polarity reversal augmentation to the original partitioned SHF to expand the dataset to 2,552 and 1,096 samples for training and testing, respectively. Similar to the first task, we replace the prediction head with a randomly initialized layer, use a batch size of 16, a RAdam optimizer with a learning rate of 5e-4, and an L1 loss between the output and the mean velocity. The model converged in 84 epochs taking 22 minutes.

\begin{figure}[h]
    \centering
    \subfloat[]{\includegraphics[height=6cm]{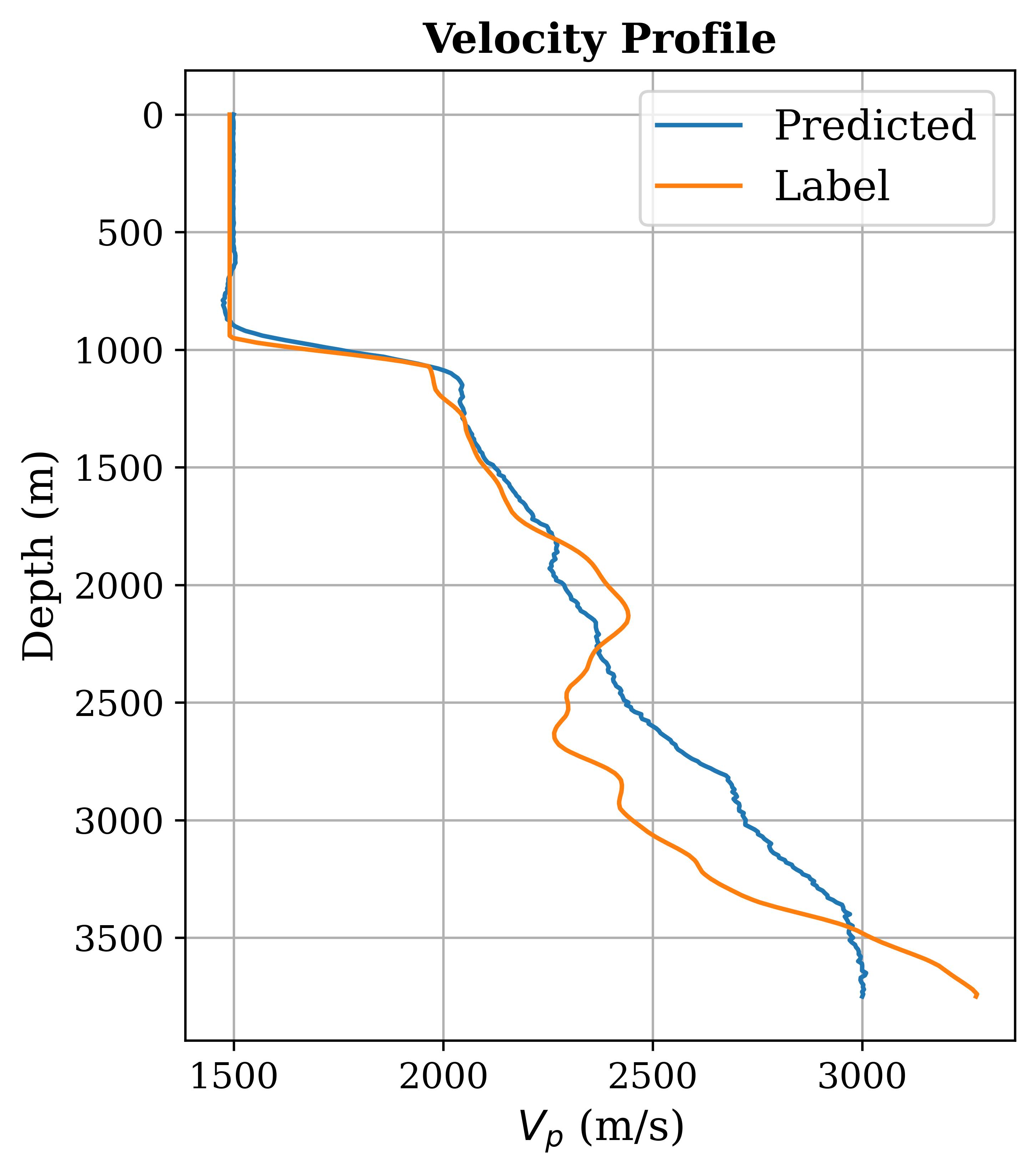} \label{fig:fig10a}}
    \subfloat[]{\includegraphics[height=6cm]{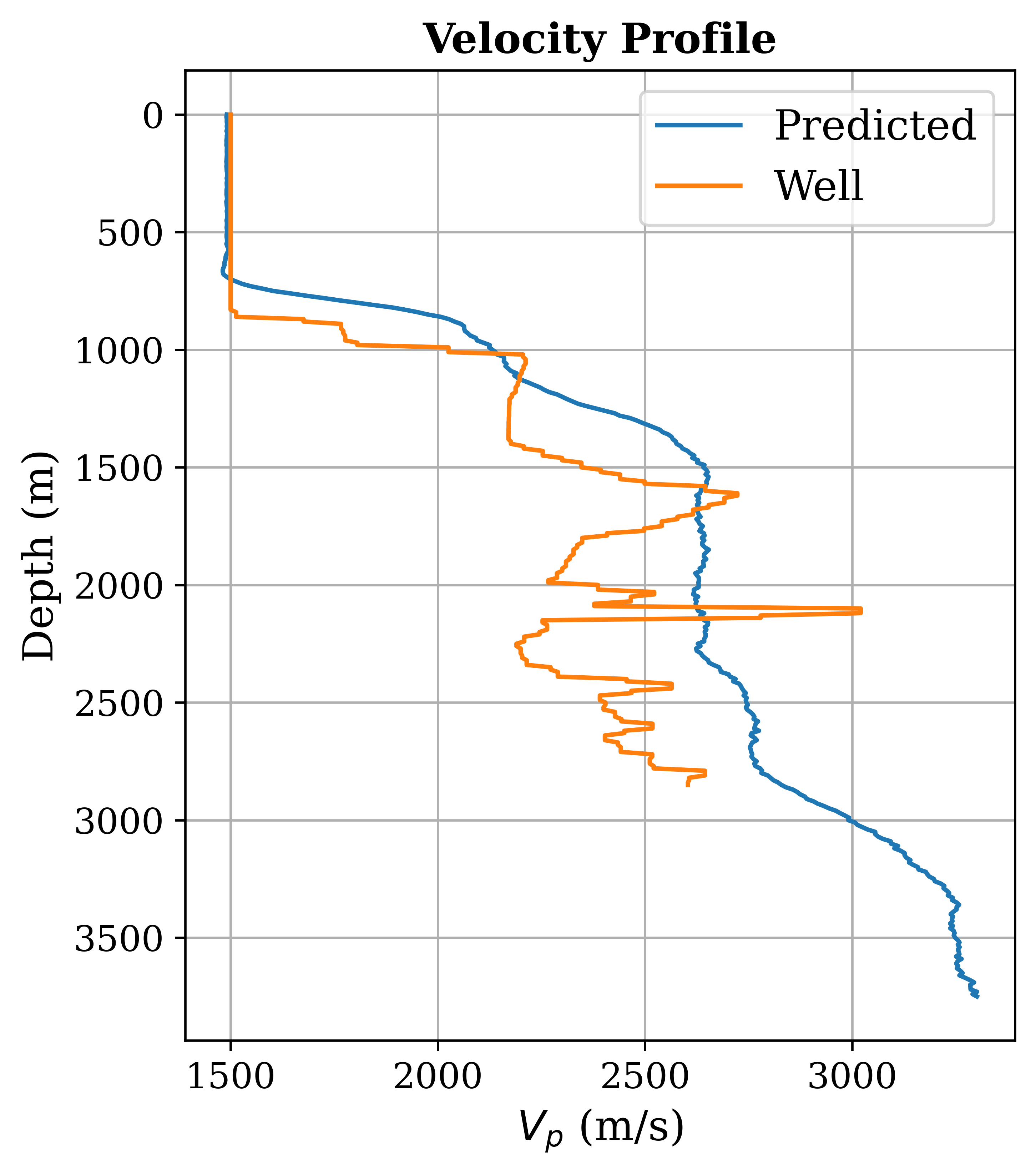} \label{fig:fig10b}}
    \caption{Examples of velocity estimation using the fine-tuned model on synthetic data (SHF, a) and field data (FHF, b). The input correspond to SHF and FHF shown in Figure \ref{fig:fig8}. Blue line represents the predicted velocity and orange line represents the label.}
    \label{fig:fig10}
\end{figure}

We achieved a test MAE of 182.7 m/s with an example of the result on SHF shown in Figure \ref{fig:fig10a}. The values of the velocity are estimated reasonably well, and the overall trend is captured in the prediction result, considering that a shot gather was used to estimate this velocity. Using a well along the field data line, we input an FHF at the well location, and compare the prediction with the well velocity, depicted in Figure \ref{fig:fig10b}. Although the details could not be captured accurately, the model output reproduced the general trend of the $V_p$ log data. Some of the difference can be attributed to anisotropy, known to be present, but ignored in the synthetic training data \cite{ovcharenko2021data}. In this case, the predicted interval (NMO-like) velocity will be higher in average than the well vertical velocity \cite{alkhalifah2016research}.

\subsection{First-break picking}
\label{sec:firstbreak_field}
First-break picks are commonly used for static correction or near surface tomography (e.g. \cite{waheed2021pinntomo}). We use the original partitioned SHF and apply a polarity reversal augmentation to expand the dataset into 2,552 and 1,096 samples for training and testing, respectively, and use them as the input. To prepare the labels, we calculate numerically the first arrivals using an eikonal solver \footnote{\url{https://github.com/scikit-fmm/scikit-fmm}} for the same random models used to generate the synthetic data, and assign the time sample index for the first arrival at each offset of the corresponding shot gather. We construct the problem as a classification task in which the model predicts the time sample index of the first-arrival at each offset from 376 indices given a shot gather as the input. For that, we replace the prediction head with a sigmoid activation function followed by a linear layer, use cross entropy loss between the output and the label, and the same batch size and optimizer used in the previous two tasks. The training took 14 epochs and 7 minutes to converge.

\begin{figure}[h]
    \centering
    \subfloat[]{\includegraphics[width=5cm]{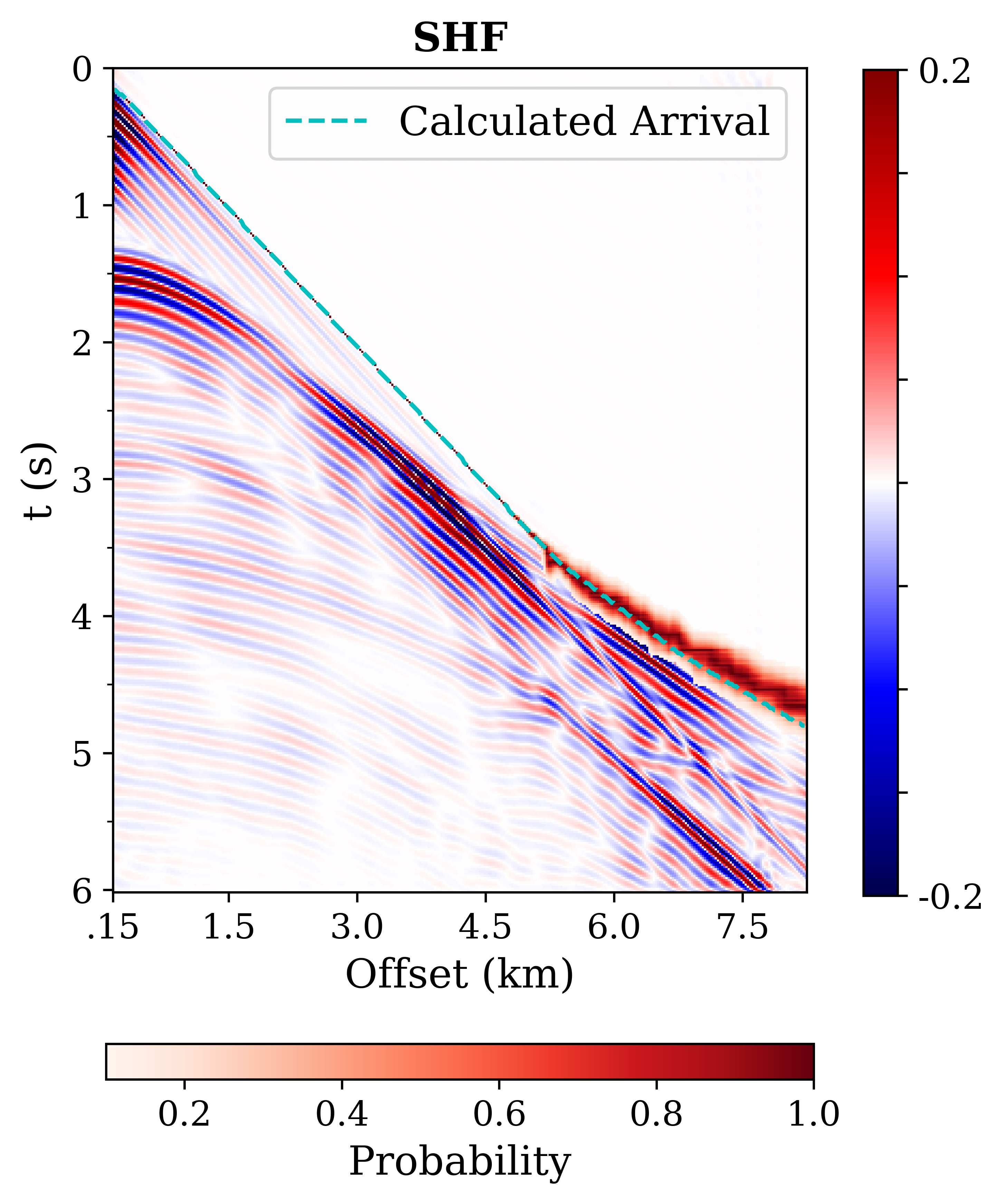} \label{fig:fig11a}}
    \subfloat[]{\includegraphics[width=5cm]{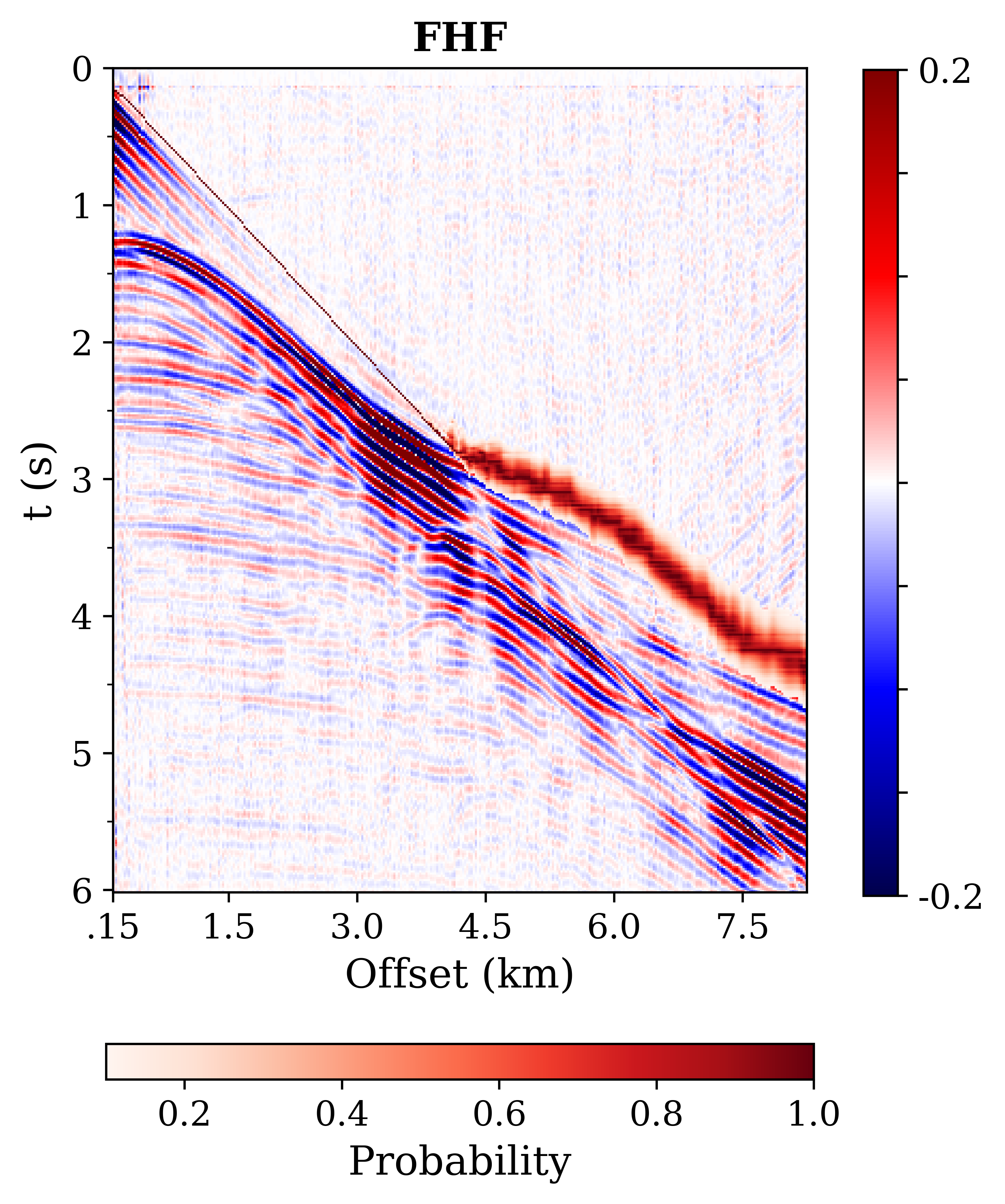} \label{fig:fig11b}}
    \caption{Examples of first-break picking using the fine-tuned model applied here on test synthetic data (SHF, a) and field data (FHF, b) samples. The input correspond to SHF and FHF shown in Figure \ref{fig:fig8}. Dashed cyan line is the first arrivals calculated numerically through eikonal solver. The superimposed red intensity marks at each pixel reflect the probability of the corresponding pixel being the first arrival.}
    \label{fig:fig11}
\end{figure}

We achieve an accuracy of the prediction of 0.75 in the test set, with a threshold of probability of 0.5. An example of the prediction for SHF is shown in Figure \ref{fig:fig11a}. For the near offset, the model produced an accurate prediction of the arrivals with small variances. However, as the offset increases, the uncertainty increases, which can be attributed to the lower amplitude at far offsets, though the true arrivals still falls within the range of the uncertainty. This good performance could also be observed in the inference on field data, the FHF example (Figure \ref{fig:fig11b}). The prediction result is quite good enough to be utilized directly or as a guide to manual picking.

\subsection{V\textsubscript{rms} prediction}
The root-mean-square (RMS) velocities, or simply $V_{rms}$, are often measured directed (and approximately) from the data, which sorted in common midpoint gathers, as normal moveout (NMO) velocities. Their values are typically useful in NMO correction to stack that data (reduce their dimension), and they can be used to extract the interval velocity for simple layered media. Specifically, the $V_{rms}$ is defined as:

\begin{equation}
    \label{eq:eq10}
    {V_{rms}}_N = \sqrt{\frac{\sum_{i=1}^{N} V_i^2 \Delta t_i}{\sum_{i=1}^{N} \Delta t_i}}
\end{equation}

where $V_i$ is the velocity in $i$-th layer and $\Delta t_i$ is the two-way vertical travel time in that layer. Using the interval velocities of the synthetic models, we calculate the corresponding RMS velocities using Equation \ref{eq:eq10}. Similar to what we did in Section \ref{sec:velpred_field}, we average the $V_{rms}$ laterally for each shot gather up to the half offset and use them as the labels. We again use a polarity reversal augmentation to the original partitioned SHF to expand the dataset to 2,552 and 1,096 samples for training and testing, respectively. Then, we replace the prediction head with a randomly initialized layer, and fine-tune using a batch size of 16, a RAdam optimizer with a learning rate of 5e-4, and an L1 loss between the output and the averaged $V_{rms}$. The fine-tuning process took 60 epochs for 18 minutes.

\begin{figure}[h]
    \centering
    \subfloat[]{\includegraphics[height=6cm]{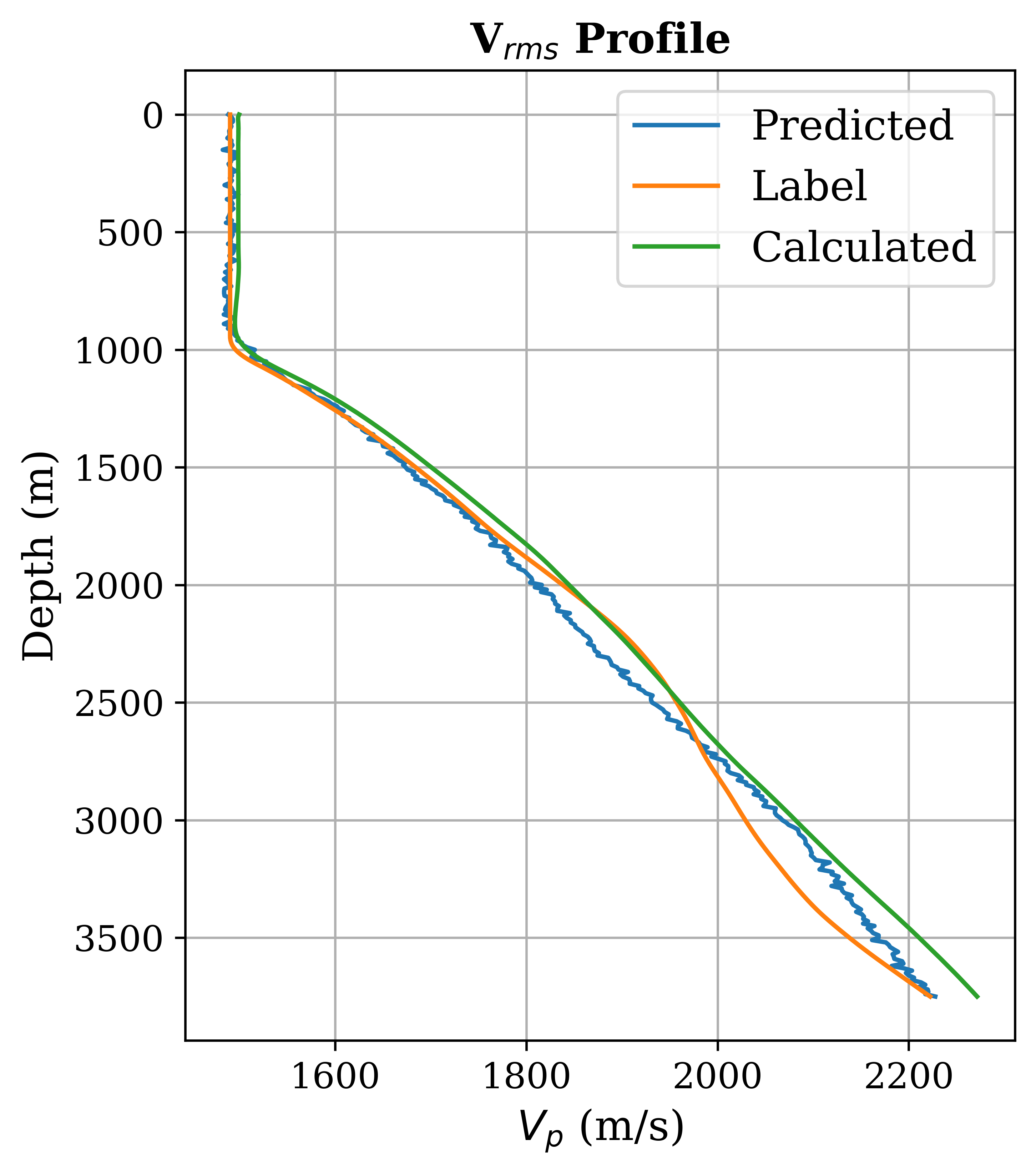} \label{fig:fig12a}}
    \subfloat[]{\includegraphics[height=6cm]{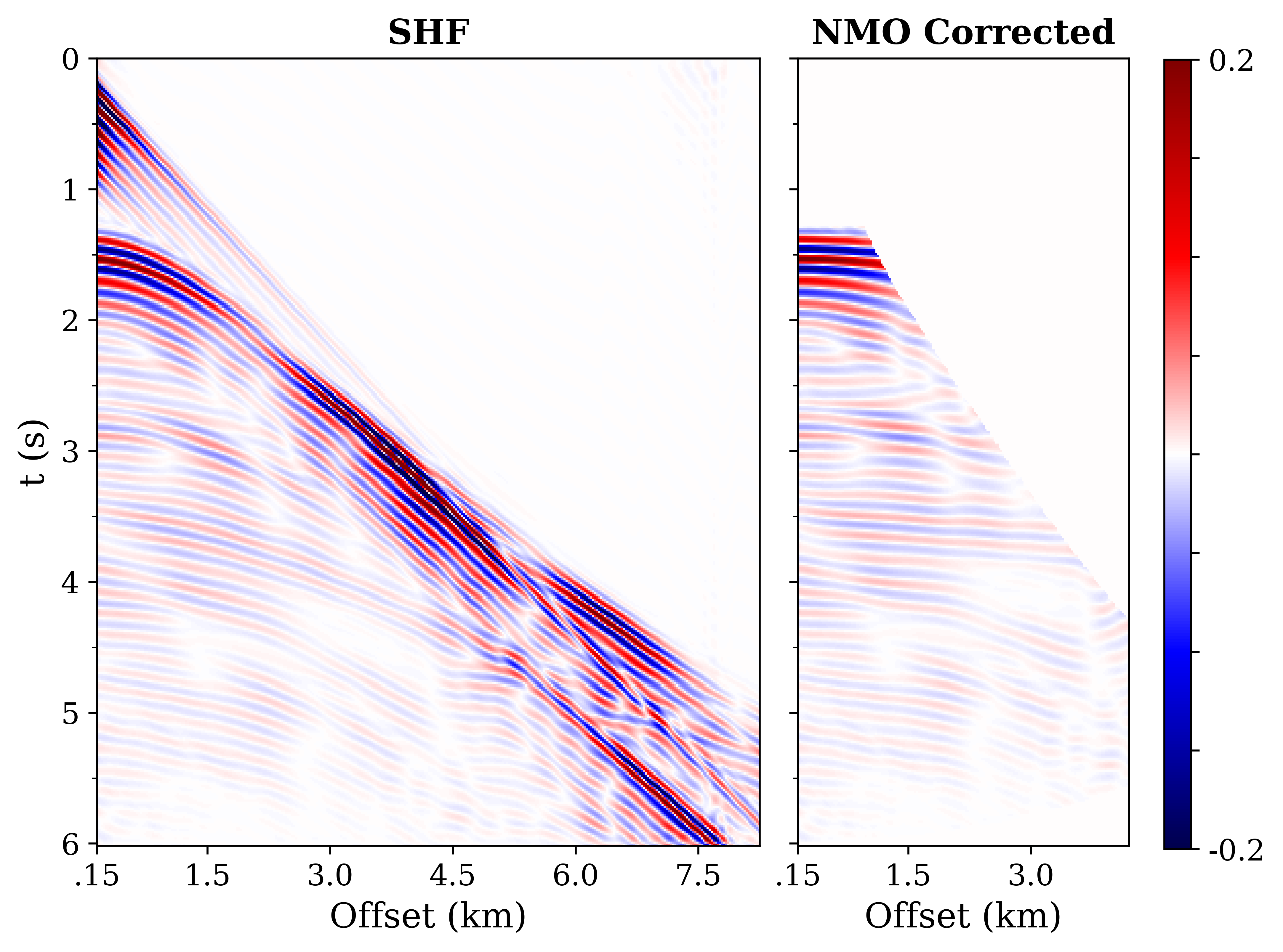} \label{fig:fig12b}}
    \caption{a) An example of $V_{rms}$ prediction on test synthetic data. Blue line represents the predicted $V_{rms}$, orange line represents the label, and green line represents the $V_{rms}$ calculated from the predicted velocity shown in Figure \ref{fig:fig10a}. b) The input synthetic shot gather (left) and the same data after NMO-correction with the predicted $V_{rms}$ (right).}
    \label{fig:fig12}
\end{figure}

\begin{figure}[h]
    \centering
    \subfloat[]{\includegraphics[height=6cm]{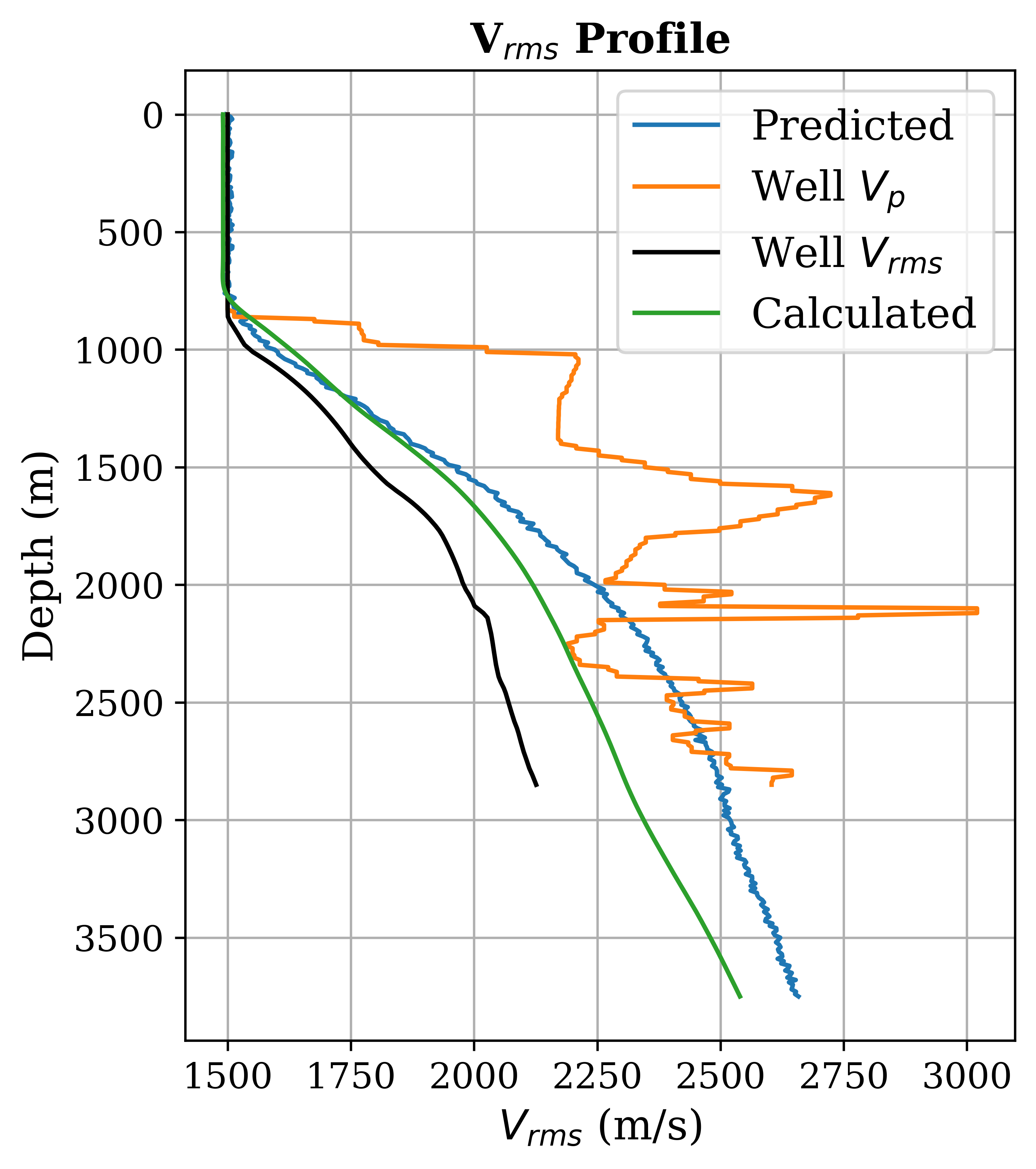} \label{fig:fig13a}}
    \subfloat[]{\includegraphics[height=6cm]{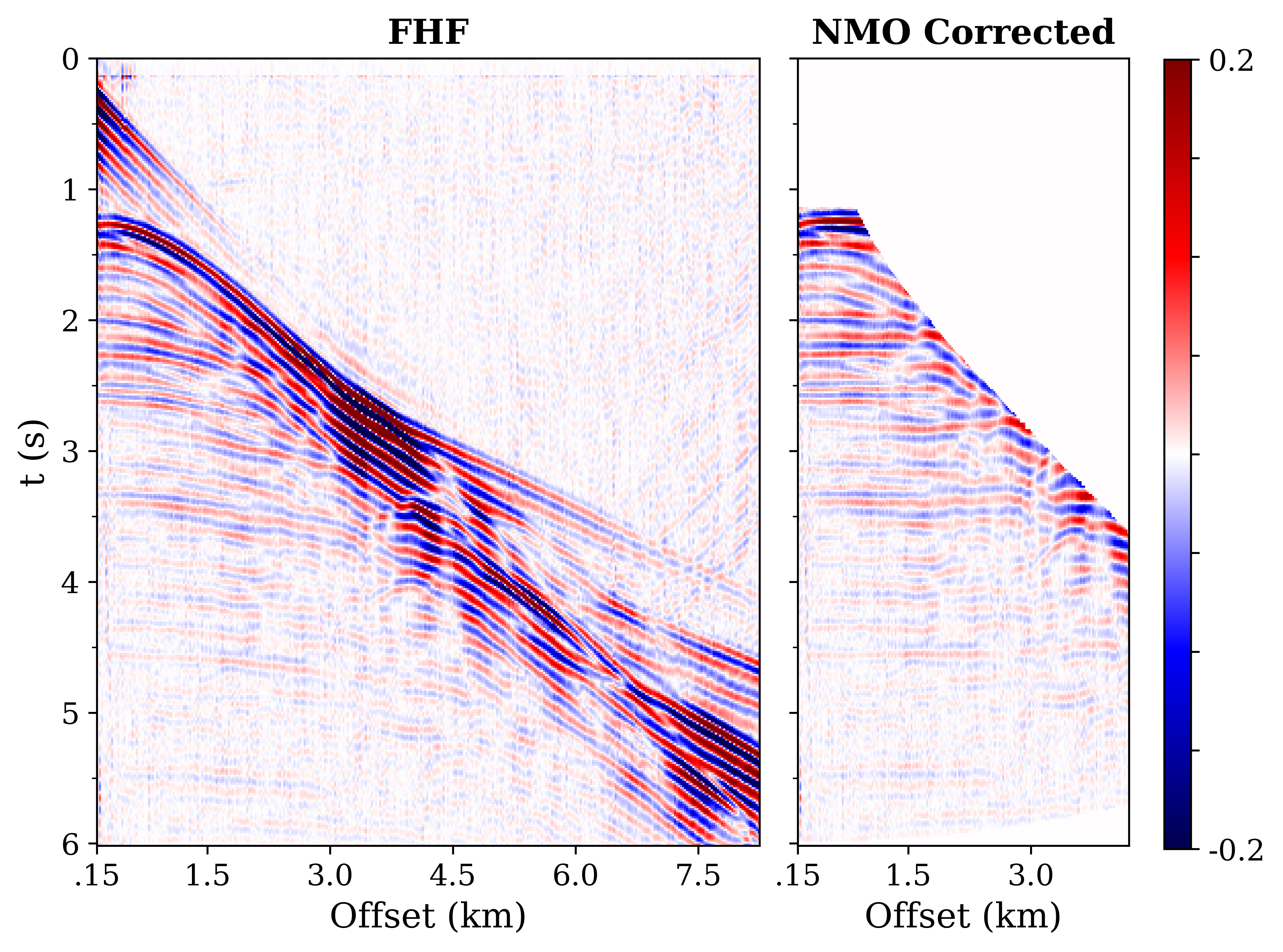} \label{fig:fig13b}}
    \caption{a) An example of $V_{rms}$ prediction on field data. Blue line represents the predicted $V_{rms}$, orange line represents the well $V_p$ measurement (similar as in Figure \ref{fig:fig10b}), black line represents the $V_{rms}$ calculated from the well $V_p$ measurement, and green line represents the $V_{rms}$ calculated from the predicted velocity shown in Figure \ref{fig:fig10b}. b) The input field shot gather (left) and the same data after NMO-correction with the predicted $V_{rms}$ (right).}
    \label{fig:fig13}
\end{figure}

We achieve an MAE of 82.2 m/s on the test set. An example of a prediction result from the synthetic test set is shown in Figure \ref{fig:fig12a}, which shows a good fit with the label. We use the predicted $V_{rms}$ to perform NMO correction to the corresponding data up to 50\% of the offsets, shown in Figure \ref{fig:fig12b}, and apply muting to avoid the NMO stretch. The "hockey sticks" effect at deeper events at the edge may be attributed to the structure (i.e. dip) of the reflectors, in which we ignore when calculating the $V_{nmo}$ (moveout velocity). Nevertheless, these automatic NMO velocity picks can be used as guidance for the manual, often more accurate, picks. We repeat the same procedure for the field data, depicted in Figure \ref{fig:fig13a}, which shows similar performance. For comparison, we calculate the $V_{rms}$ from the velocity prediction result shown in Figure \ref{fig:fig10} (Figure \ref{fig:fig12a} and \ref{fig:fig13a}), and we observe reasonable consistency, especially the trend. Like for the interval velocities (Section \ref{sec:velpred_field}), we attribute the difference to the presence of anisotropy not taken into account in the synthetic data fine tuning, in which the well velocities are lower than the seismic velocities (the classic seismic misties issue \cite{li2018using}). The interesting observation here is the better flattening of the field data after NMO correction with the predicted $V_{rms}$ (Figure \ref{fig:fig13b}), as compared to the synthetic data (Figure \ref{fig:fig12b}). This can potentially be attributed to the reduced lateral inhomogeneity of the subsurface at this location compared to the generated models for the synthetic data (Figure 7).

\section{Discussion}
\label{sec:discussion}
In this section, we will share our experience on key model and training variables for StorSeismic including the model architecture, the field/synthetic data ratio in pre-training, and the degree of fine-tuning needed to achieve good inference results. In other words, which layers we need to fine-tune.

\subsection{Model configuration}
\begin{figure}[!h]
    \centering
    \includegraphics[width=0.8\textwidth]{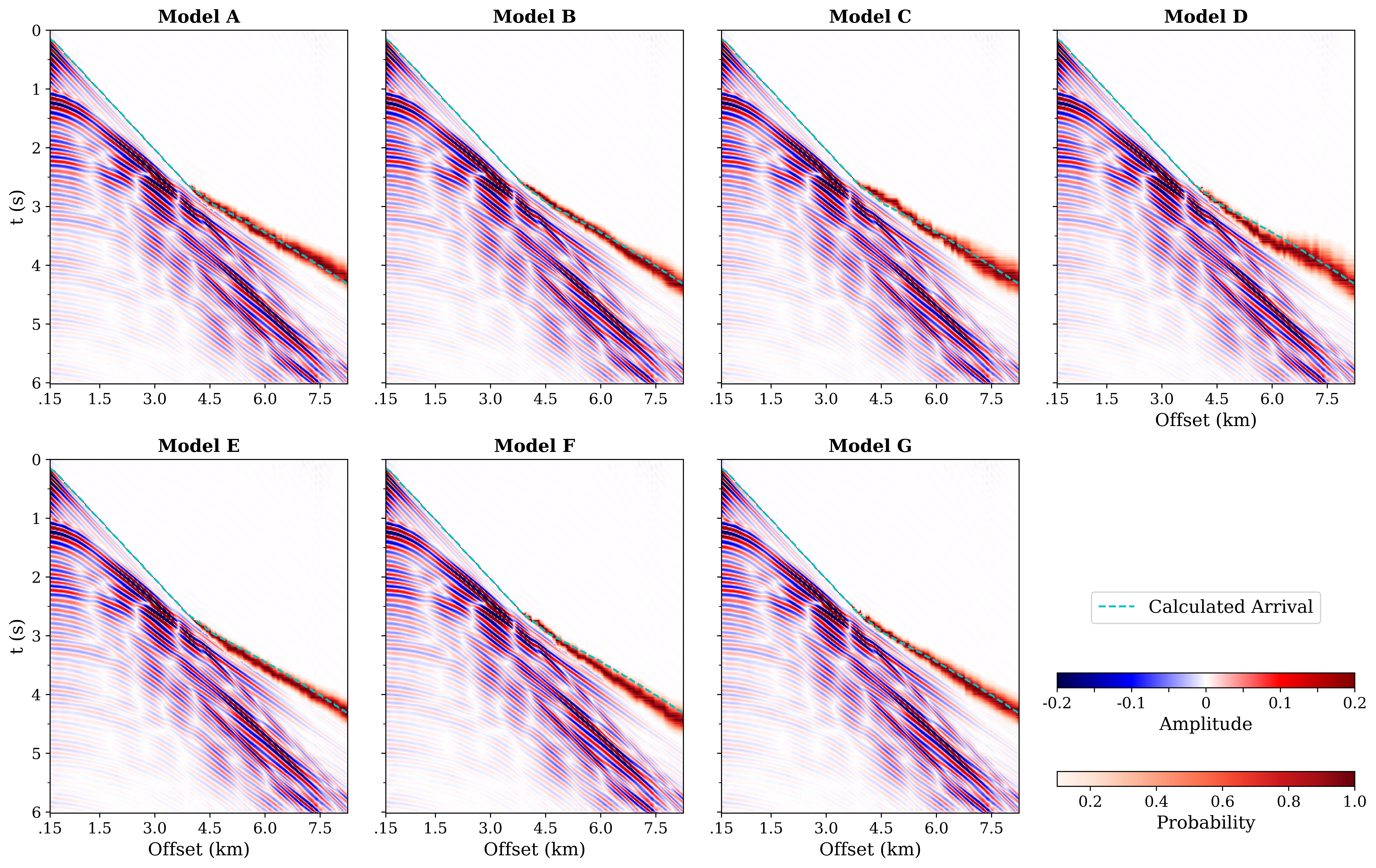}
    \caption{Prediction results of the first-break picking task on a synthetic test example for different model configurations.}
    \label{fig:fig17}
\end{figure}

\begin{figure}[!h]
    \centering
    \includegraphics[width=0.8\textwidth]{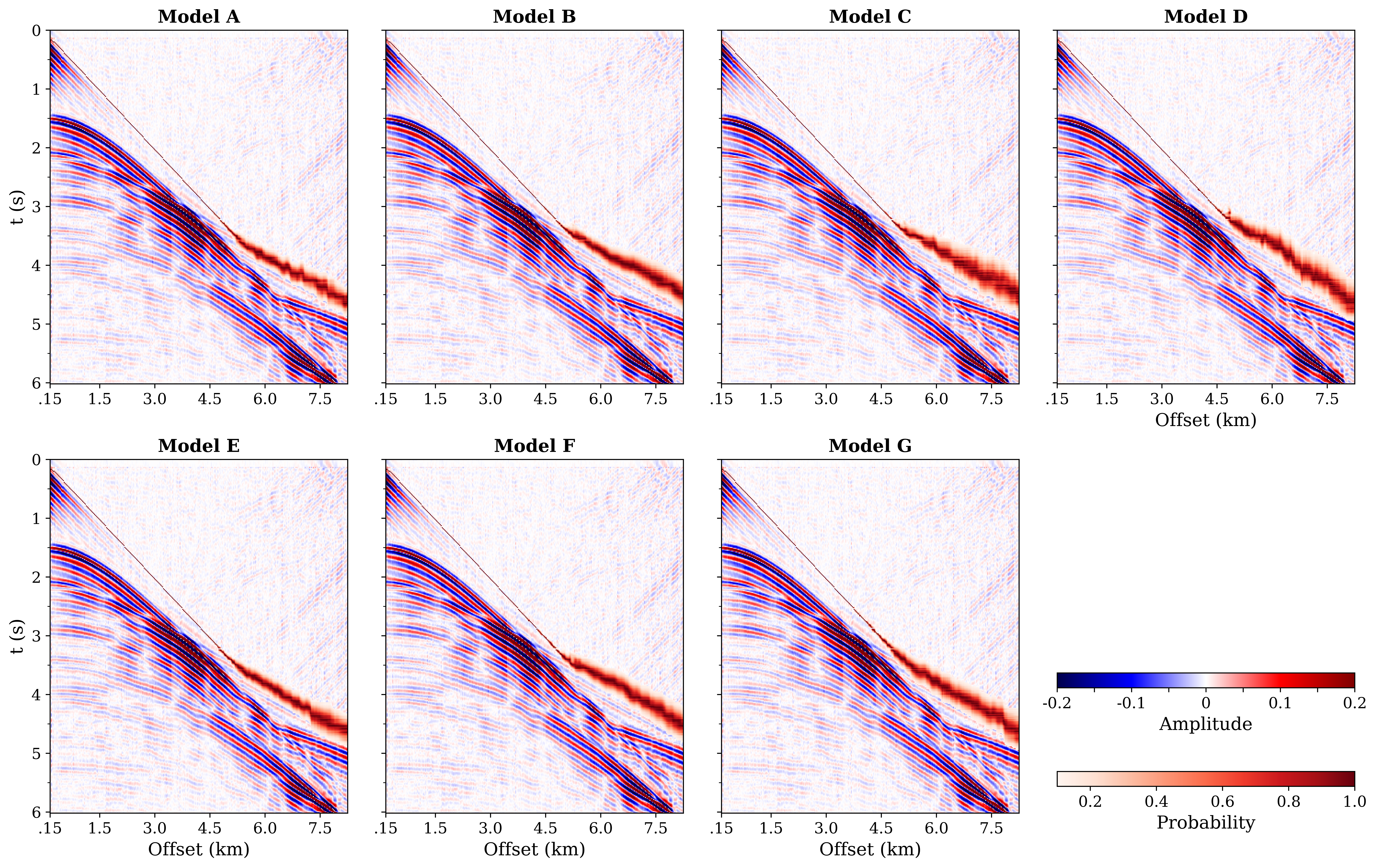}
    \caption{Prediction results of the first-break picking task on a field data example for different model configurations.}
    \label{fig:fig18}
\end{figure}

The BERT architecture came in different sizes and configurations when it was first proposed \cite{devlin2018bert}. In this study, we choose to use BERT\textsubscript{MINI} (H = 256, L = 4, A = 4) as the baseline configuration, adhering to the suggestion that Transformers are universal computational functions \cite{lu2021pretrained}. However, for training cost consideration, we explore how much do we gain and/or lose when we increase or decrease the hidden dimension (H), the number of layers (L), and the number of attention heads (A). We perform this test on the first-break picking task, all with the same pre-training setup as in Section \ref{sec:pretrain_field} and the fine-tuning setup described in Section \ref{sec:firstbreak_field}, except that we did not freeze any attention layers (see Section \ref{sec:technique}). The summary of the results of this test is shown in Table \ref{tab:table1}, with model A being the model configuration that we use in this study. An example of the comparison on the synthetic data is shown in Figure \ref{fig:fig17}, and that of the field data is shown in Figure \ref{fig:fig18}.

\label{sec:configuration}
\begin{table}
 \caption{Various model configurations, the training cost, and the results. Bold numbers correspond to the optimal values. The accuracy shown are calculated using a threshold of probability of 0.5.}
  \centering
  \begin{tabular}{ccccccccccc}
    \toprule
    % \multicolumn{5}{c}{} & \multicolumn{3}{c}{Pre-training} & \multicolumn{3}{c}{Fine-tuning (first-break)} \\
    % \cmidrule(r){6-8} \cmidrule(r){9-11}
    % Model & H & L & A & Params. & Epoch & Time & Test loss & Epoch & Time & Accuracy \\
    \multirow{2}{*}{Model} &
    \multirow{2}{*}{H} &
    \multirow{2}{*}{L} &
    \multirow{2}{*}{A} &
    \multirow{2}{*}{Params.} &
    \multicolumn{3}{c}{Pre-training} & \multicolumn{3}{c}{Fine-tuning (first-break)} \\
    \cmidrule(r){6-8} \cmidrule(r){9-11}
    & & & & & Epoch & Time & Test loss & Epoch & Time & Accuracy \\
    \midrule
    A & 256 & 4 & 4 & 3,352,696 & 267 & 12.7h & 1.74e-5 & 15 & 7.3m & 0.72 \\
    B & 128 & 4 & 4 & \textbf{890,104} & 197 & 8.5h & 3.41e-5 & 26 & 7.8m & 0.74 \\
    C & 512 & 4 & 4 & 12,996,472 & 257 & 17.7h & \textbf{1.16e-5} & 11 & 10.3m & \textbf{0.76} \\
    D & 256 & 2 & 4 & 1,773,176 & 232 & 10.8h & 2.62e-5 & 12 & \textbf{5.5m} & 0.75 \\
    E & 256 & 8 & 4 & 6,511,736 & 195 & 14.4h & 1.53e-5 & 16 & 10.5m & 0.72 \\
    F & 256 & 4 & 2 & 3,352,696 & 208 & 10.1h & 2.02e-5 & 16 & 8.3m & 0.73 \\
    G & 256 & 4 & 8 & 3,352,696 & 160 & \textbf{8.1h} & 1.97e-5 & 11 & 8.1m & 0.72 \\
    \bottomrule
  \end{tabular}
  \label{tab:table1}
\end{table}

Directly looking at the accuracy, we observe that there is no significant difference between all the models tested. The largest model with nearly 13M parameters, model C, has the highest accuracy of 0.76 compared to all other models. This is also reflected in the lowest pre-training loss of model C, which shows that the model learned a better representation of the seismic data. However, this comes at the cost of doubling the training time compared with the smallest model (model B), with a decrease in H, which has a close value of accuracy of 0.74. A similar performance with model A could be observed in model E which has more layers. The higher accuracy of the larger model might not translate to higher accuracy at the inference on the field data. Interestingly, increasing the number of attention heads A (model G) reduces the total training time of pre-training and fine-tuning. This is attributed to the design of the BERT architecture in which the calculation of the attention heads becomes cheaper. All in all, the performance of the proposed StorSeismic model are almost size-agnostic (to the given dataset), with the selection of the model size has to be considered when the available computational resources is limited.

\subsection{Mixing of field and synthetic data}
\label{sec:mixing}
Due to the complexity of the Earth subsurface, the currently available forward modeling software are only able to handle this complexity to a certain limit, hence, it could not fully represent seismic waves propagating through real Earth media. It is then advantageous to account for this limitation, in the proposed framework, by including the field data as part of the pre-training dataset. Thus, here, we test what proportion of the field data is actually needed in the pre-training and how much of would impact the inference stage. We will use the velocity estimation task for this test. For this purpose, we formulate two other versions of the StorSeismic model: 1). Pre-trained using dataset with 15\% of which are field data, and 2). Pre-trained using 30\% field data, and compare them with the 50\% presented in Section \ref{sec:velpred_field}. We preserve the total number of pre-training dataset by generating more synthetic data from the same random subsurface models as in Section \ref{sec:examples}, and use the same training setup as in Section \ref{sec:pretrain_field} for all. In the fine-tuning, we use the same dataset and training setup as in Section \ref{sec:velpred_field} for all experiments.

\begin{figure}[h]
    \centering
    \subfloat[]{\includegraphics[width=5.5cm]{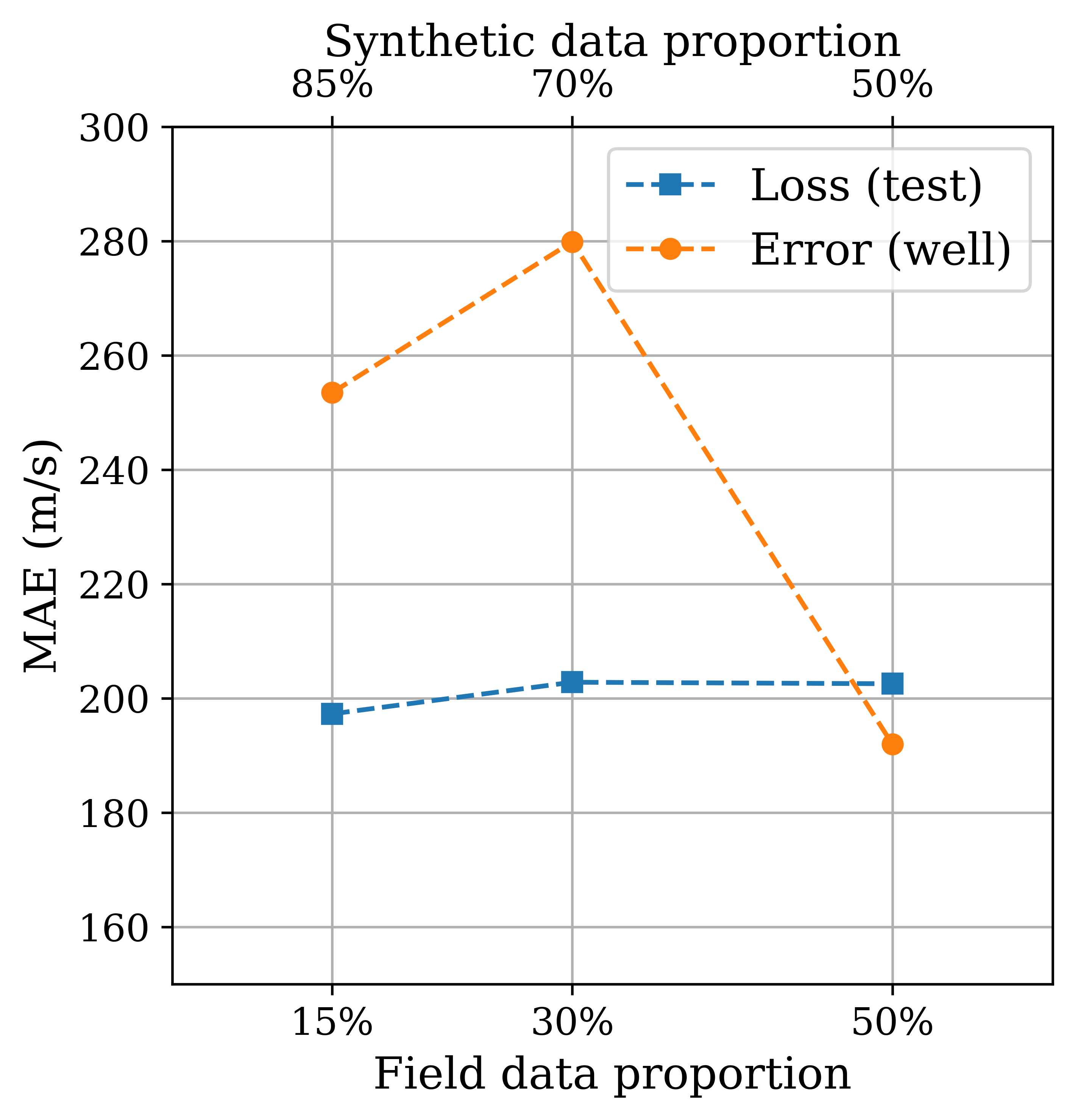} \label{fig:fig16a}}
    \subfloat[]{\includegraphics[height=6cm]{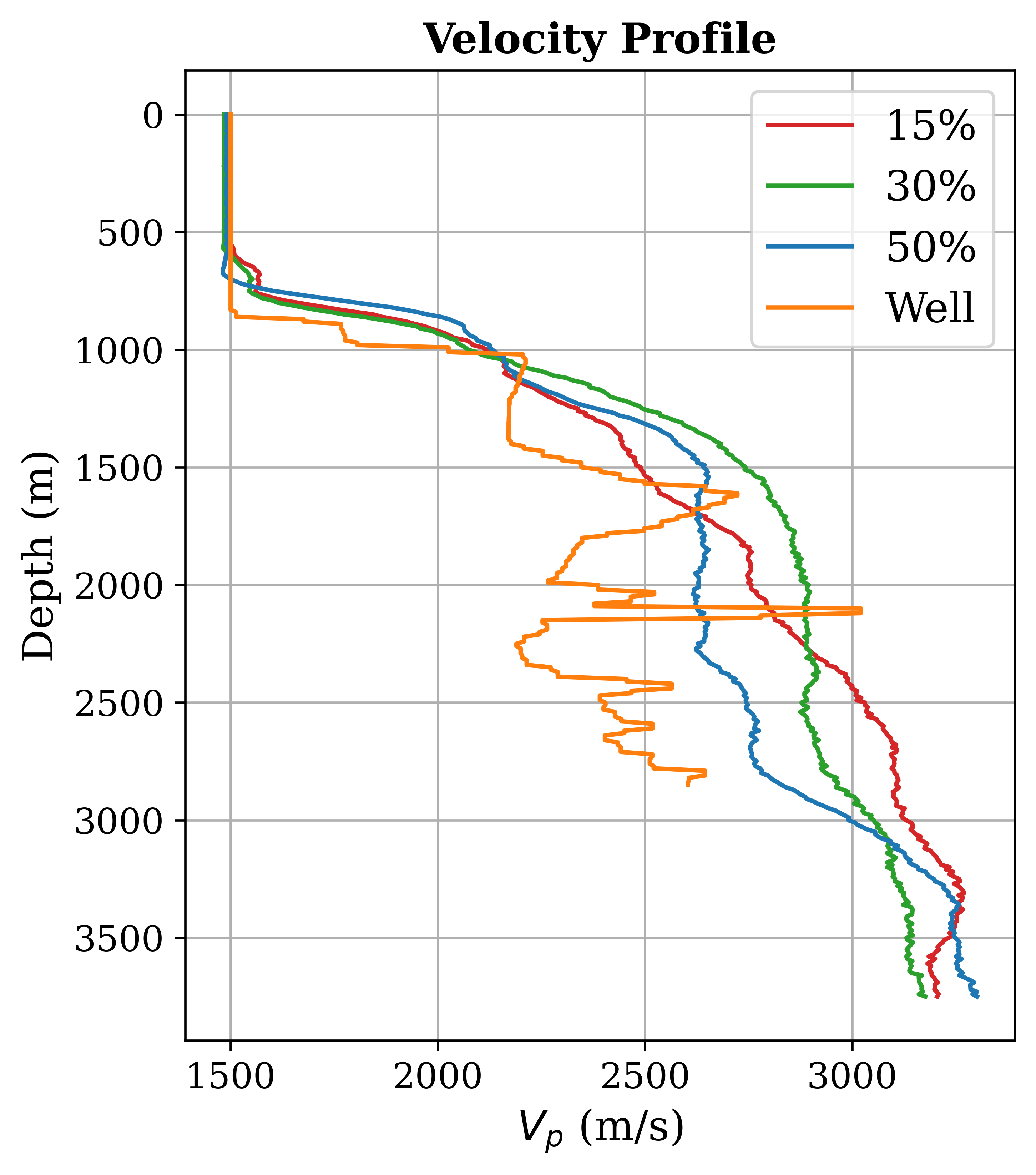} \label{fig:fig16b}}
    \caption{a) Fine-tuning loss on synthetic test set (blue dots) and the error of the prediction on field data with a well data (orange dots) for different proportions of field data used in the pre-training. b) Comparison of the predictions with the well data. The percentage on the legend represents the proportion of the field data used in the pre-training.}
    \label{fig:fig16}
\end{figure}

A reliable source of information to compare with from our field data is well log measurements of the velocity (Figure \ref{fig:fig10b}). Therefore, we treat the $V_p$ log data (check-shot velocity) as the ground truth and measure the error with the prediction at the well location from all versions of the model, along with their corresponding loss on the fine-tuning test set, shown in Figure \ref{fig:fig16a}. We observe an increasing trend of test loss as the proportion of the field data increases (i.e. decreases in the synthetic data proportion). Conversely, the error of the prediction with the well data decreases significantly as the field data proportion increases, which reflects in the results shown in Figure \ref{fig:fig16b}. The reduced loss for the inference suggests the importance of having a reasonable amount of field data in the pre-training so its features are also stored. On the other hand, we have to include a good proportion of the synthetic data in the pre-training as we fine tune with synthetic data. We plan to dedicate additional research to understanding the optimal proportion of each data and optimal amount of data in general needed for pre-training. However, the proportions we used in this study seems to admit credible results.

\subsection{Fine-tuning options}
\label{sec:technique}

As \cite{merchant2020happens} suggested, changes in BERT's representation mainly occur at the deeper layers (i.e., layers that are closer to the output). We also observe this in our results in Section \ref{sec:analysis}. Therefore, following \cite{merchant2020happens}, we test the "partial freezing" process in our experiment with the field data. Starting from the pre-trained model described in Section \ref{sec:pretrain_field}, we fine-tune for the denoising task (Section \ref{sec:denoising_field}) by freezing the first attention layer, first 2 layers, first 3 layers, and compare them with no freezing (i.e. fine-tuning all attention layers), all with the same training (fine-tuning) configuration described in Section \ref{sec:denoising_field}, but, here, we allow them train to convergence using an early stopping scheduler as the limiter.

\begin{figure}[!h]
    \centering
    \includegraphics[width=8cm]{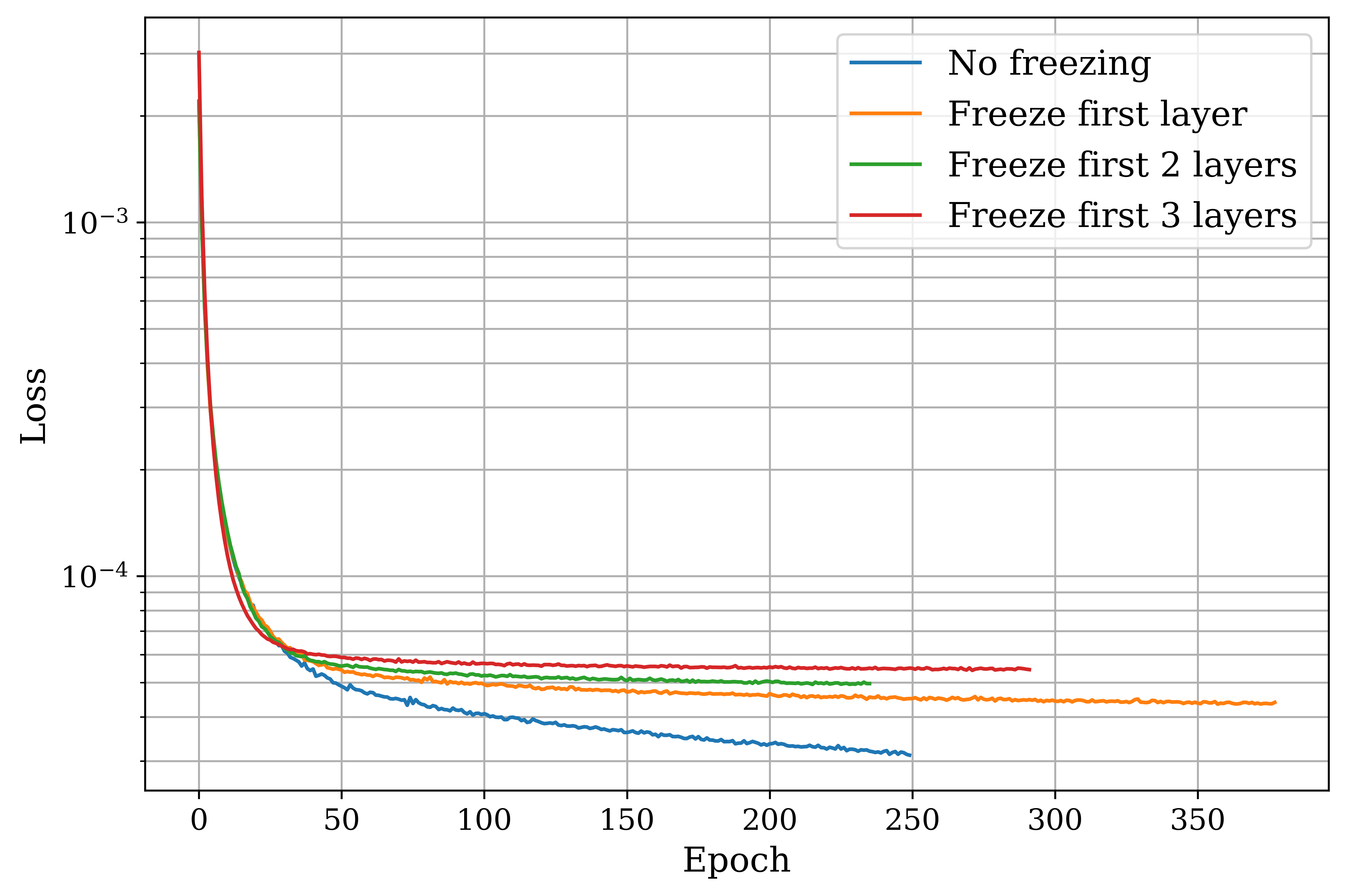}
    \caption{The fine-tuning test loss curve for the denoising task for different numbers of frozen attention layers.}
    \label{fig:fig14}
\end{figure}

\begin{figure}[!h]
    \centering
    \includegraphics[width=1\textwidth]{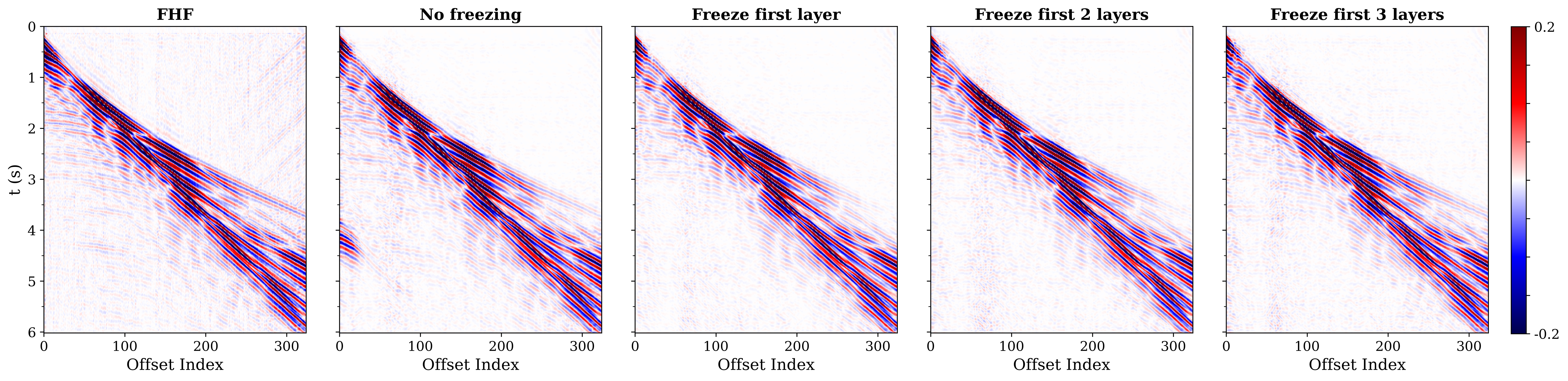}
    \caption{Examples of denoising results on the field data for different numbers of frozen attention layers.}
    \label{fig:fig15}
\end{figure}

The fine-tuning test loss curve shown in Figure \ref{fig:fig14} implies that the freezing options lead to different minima. It also suggests that the loss decreases as the degrees of freedom increases (i.e., the number of frozen layers decrease). However, the inference on the field data is our main goal. Thus, we share the predictions for the various fine-tuning options on the field data in Figure \ref{fig:fig15}. Unfortunately, fine-tuning all layers leads to unwanted signal in the bottom-left part of the denoised shot gather, which can be attributed to the fact that the network adapted too much to the synthetic data, possibly overriding most of the field data features learned in the pre-training. Though, we note that we also lose some signal of the deeper events as the number of frozen layers increase, we conclude that it is important to freeze some layers during fine-tuning as a way to preserve the field data imprint that was learned in the pre-training. Thus, for all the fine-tuning tasks presented, we chose to freeze the first two layers of our pre-trained model as a compromise between accuracy and training time.

\section{Conclusion}
\label{sec:conclusion}
We introduced a new paradigm for seismic processing based on deep learning, StorSeismic, which offers an integrated framework to deal with seismic datasets to facilitate efficient and accurate processing. With the help of the self-attention mechanism in the Transformer encoder architecture, StorSeismic could capture and store the features embedded in the seismic data in the pre-training stage, and with limited change to the pre-trained model (i.e. the prediction head), we can fine-tune the network to adapt to many processing tasks. In the examples, we demonstrated the ability of StorSeismic to denoise, predict velocity models, pick first arrivals, and produce NMO velocities.

The proposed framework also allows us to address the problem of dealing with label-less field data. We utilized the self-supervised nature of the pre-training, which is tasked with storing seismic data features, to store both the field and synthetic data features, which will consequently increase the model's capacity to handle field data in the inference stage.

\section*{Acknowledgments}
We would like to thank Bingbing Sun for his initial work on this concept. We thank KAUST for its continuous support and SWAG, especially Claire Birnie, for the fruitful discussions. We thank CGG for the field seismic data and Oleg Ovcharenko for his help in retrieving the data.

%Bibliography
% \nocite{*}
\bibliographystyle{unsrt}  
\bibliography{references}

\end{document}